\documentclass[10pt,twocolumn,letterpaper]{article}

\usepackage{cvpr}              

\usepackage{adjustbox}
\usepackage{algorithm}
\usepackage{algorithmic}

\usepackage{color}
\usepackage{comment}
\usepackage{enumerate}
\usepackage{multirow}
\usepackage{subcaption}
\usepackage{bbm}

\usepackage{wrapfig}
\usepackage{xcolor}         
\usepackage[table]{xcolor}  
\usepackage{pifont}
\usepackage{tcolorbox}  
\usepackage{tikz}
\usetikzlibrary{arrows.meta, positioning}
\usepackage{inconsolata}

\newcommand{\StateSet}{\mathcal{S}}
\newcommand{\ActionSet}{\mathcal{A}}
\newcommand{\Dynamics}{\mathcal{T}}

\newcommand{\instr}{\ell}
\newcommand{\Demo}{\mathcal{D}}
\newcommand{\cmark}{\textcolor{green!60!black}{\ding{51}}}
\newcommand{\xmark}{\textcolor{red}{\ding{55}}}

\DeclareMathOperator*{\argmax}{argmax}

\newcommand{\ours}{\textsc{NeSyCR}}

\usepackage{listings}
\tcbuselibrary{breakable}
\usepackage{enumitem}

\definecolor{cvprblue}{rgb}{0.21,0.49,0.74}
\usepackage[pagebackref,breaklinks,colorlinks,allcolors=cvprblue]{hyperref}
\usepackage[accsupp]{axessibility} 

\def\paperID{XXXXX} 
\def\confName{CVPR}
\def\confYear{2026}

\title{Cross-Domain Demo-to-Code via Neurosymbolic Counterfactual Reasoning}

\author{Jooyoung Kim, Wonje Choi, Younguk Song, Honguk Woo\thanks{Honguk Woo is the corresponding author.}\\
Department of Computer Science and Engineering, Sungkyunkwan University\\
{\tt\small \{onsaemiro, wjchoi1995, syw2045, hwoo\}@skku.edu}
}

\begin{document}
\maketitle
\addtocontents{toc}{\protect\setcounter{tocdepth}{-10}}

\begin{abstract}
Recent advances in Vision-Language Models (VLMs) have enabled video-instructed robotic programming, allowing agents to interpret video demonstrations and generate executable control code.
We formulate video-instructed robotic programming as a cross-domain adaptation problem, where perceptual and physical differences between demonstration and deployment induce procedural mismatches. However, current VLMs lack the procedural understanding needed to reformulate causal dependencies and achieve task-compatible behavior under such domain shifts.
We introduce $\ours$, a neurosymbolic counterfactual reasoning framework that enables verifiable adaptation of task procedures,  providing a reliable synthesis of code policies. 
$\ours$ abstracts video demonstrations into symbolic trajectories that capture the underlying task procedure. Given deployment observations, it derives counterfactual states that reveal cross-domain incompatibilities. By exploring the symbolic state space with verifiable checks, $\ours$ proposes procedural revisions that restore compatibility with the demonstrated procedure.
$\ours$ achieves a 31.14\% improvement in task success over the strongest baseline Statler, showing robust cross-domain adaptation across both simulated and real-world manipulation tasks.
\end{abstract}

\section{Introduction}\label{sec:intro}
Advances in foundation models have accelerated progress toward general-purpose embodied intelligence, enabling robots to interpret human instructions and execute complex tasks as autonomous control policies~\cite{fdagent:palme, fdagent:saycan, fdagent:inner, fdagent:llmplanner}.
In particular, Large Language Models (LLMs) with code-writing capabilities have inspired the Code-as-Policies paradigm, in which executable control code is synthesized from language instructions using predefined APIs~\citep{cap:cap, cap:chatgptforrobotics, cap:instruct2act, cap:voxposer}.
Furthermore, Vision-Language Models (VLMs) have extended this paradigm toward a general form of video-instructed robotic programming, where robotic programs are generated from instructional video demonstrations by translating observed task sequences into structured task specifications that can be compiled into control code~\cite{vlcap:demo2code, vlcap:robotic, vlcap:seedo, vlcap:video2policy}.
By capturing richer perceptual context and task intent from demonstrations, these methods enable more grounded robotic programming than text instructions alone.

\begin{figure*}[th]
\begin{center}
\includegraphics[width=0.99\linewidth]{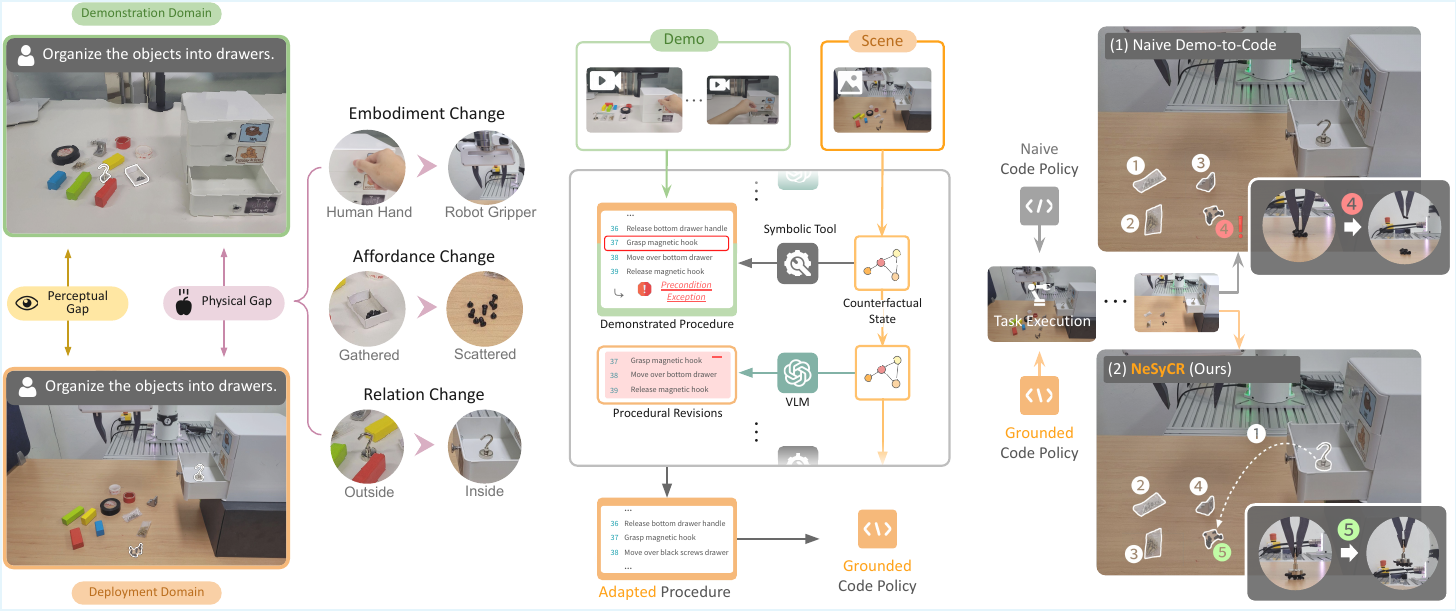}
\end{center}
\vspace{-15pt}
\caption{
Overview of $\ours$ in a drawer-organizing task scenario. \textbf{(Left)} Illustration of the domain gap between the demonstration and deployment.
\textbf{(Middle)} Overview of $\ours$ framework, which generates an adapted procedure via neurosymbolic counterfactual reasoning.
\textbf{(Right)} Outcome of the adapted procedure, showing that $\ours$ successfully executes the task via a grounded code policy.
}
\vspace{-10pt}
\label{fig:overview}
\end{figure*}

In video-instructed robotic programming, domain gaps between the demonstration and deployment are inevitable due to inherent differences in environmental layouts, object properties, and morphological constraints~\cite{LFDRL, LFO, LFD, emb:conpe}.
While sensory observations can reveal physical discrepancies between the demonstration and the deployment environment, these observations alone do not explain how structural differences disrupt the underlying task procedure or how actions will causally unfold under altered conditions. 
To address this limitation, we present $\ours$, a neurosymbolic framework that formulates cross-domain adaptation from demonstrations as counterfactual reasoning.
At its core, $\ours$ infers how task outcomes would change under altered domain factors and proposes alternative actions that revise the demonstrated behavior to restore task-oriented behavioral compatibility.
%
Given a demonstration, $\ours$ constructs a symbolic world model that encodes the task procedure as a symbolic abstraction of the trajectory. Leveraging this model, $\ours$ performs neurosymbolic counterfactual adaptation to revise the procedure for the deployment domain, enabling verifiable adaptation of the demonstrated procedure across domains.

As shown in Figure~\ref{fig:overview}, the demonstration domain involves a human performing a drawer-organizing task with objects such as a magnetic hook and a box of screws placed on the table.
In the deployment domain, however, the human hand is replaced by a robotic gripper, the magnetic hook is already arranged inside the drawer, and the screws are scattered across the surface.
%
These differences create a procedural incompatibility: direct grasping becomes infeasible, requiring the magnetic hook to be repurposed as an auxiliary tool for gathering the screws.
Furthermore, this modification introduces a cascading incompatibility: objects to be organized later in the procedure can obstruct the magnet, rendering it inaccessible.
Resolving these coupled incompatibilities requires reordering the procedure so that the magnet is retrieved and used to aggregate the screws before the target objects are placed.
However, a direct demo-to-code approach using VLM-based prompting would still attempt to grasp the scattered screws with the gripper, which fails due to their small size and dispersion.
In contrast, $\ours$ identifies these incompatibilities through VLM-based symbolic translation and symbolic forward simulation. By combining the VLM’s commonsense reasoning with symbolic verification, it produces an adapted and coherent procedure that completes the task. Further details are illustrated in Figure~\ref{fig:vis}.

To construct the symbolic world model, the VLM abstracts the demonstration into a symbolic trajectory that captures objects and their spatial and temporal relations, while the symbolic tool verifies these representations for logical consistency.
In a deployment domain, $\ours$ identifies structural variations through this symbolic world model, detecting where the demonstrated procedure fails to reach the goal state under counterfactual states derived from target-domain observations.
%
%
The VLM then proposes alternative action operators for the incompatible steps, and the symbolic tool verifies their causal validity, yielding revised procedural steps that restore compatibility.
This reasoning process produces an adapted task specification that remains logically valid while preserving the task intent of the demonstration.
The adapted specification is then compiled into a reliable, deployment-grounded code policy.

We evaluate $\ours$ on a diverse set of video-instructed robotic programming scenarios, deploying robotic agents in both simulated and real-world environments. The cross-domain setting between demonstration and deployment is characterized by domain factors that capture variations in environmental and embodiment configurations.
The scenarios consist of long-horizon manipulations involving multiple subtask types and requiring up to 116 visual-motor API calls, yielding compositional and procedurally complex tasks.
$\ours$ outperforms the strongest baseline, Statler~\cite{wm:statler}, achieving an average improvement of 31.14\% in task success rate, as reported in Tables~\ref{tab:main:sim} and~\ref{tab:main:real}.

Our contributions are summarized as follows:
(1) We present the $\ours$ framework that casts cross-domain adaptation from demonstrations as counterfactual reasoning for video-instructed robotic programming.
(2) 
We implement a VLM–symbolic tool pipeline that proposes and verifies alternative action steps, ensuring procedural compatibility across domains.
(3) We evaluate $\ours$ across simulated and real-world robotic manipulation tasks using an experimental design that provides high granularity over domain factors and task complexity, enabling precise analysis of its cross-domain procedural adaptation.

\section{Problem formulation}\label{sec:probl}
We formulate the embodied domain as a tuple $\mathcal{M}\!=\!(\StateSet,\!\ActionSet,\!\Dynamics,\!g)$, where $\StateSet$ denotes the state space, $\ActionSet$ the action space, $\Dynamics\!:\!\StateSet\!\times\!\ActionSet\!\rightarrow\!\StateSet$ the transition function, and $g\!\in\!\mathcal{G}$ the goal state.
Under partial observability~\cite{sutton2018reinforcement}, the agent receives observations $o_t\!\in\!\mathcal{O}$ at each timestep $t$ via an observation function $\Omega\!:\!\StateSet\!\times\!\ActionSet\!\rightarrow\!\mathcal{O}$, which maps the underlying state to perceptual observations (e.g., RGB images).
%
%
In cross-domain settings, a demonstration $\Demo\!=\!(\!\{o_t\}_{t=1}^N,\instr)$ specifies a domain $\mathcal{M}_{\mathrm{S}}\!=\!(\StateSet_{\mathrm{S}},\!\ActionSet_{\mathrm{S}},\!\Dynamics_{\mathrm{S}},\!g)$ performed under specific environmental and embodiment configurations, optionally with a language description $\instr$.
%
%
The agent must achieve the same goal in a deployment domain $\mathcal{M}_{\mathrm{T}}\!=\!(\StateSet_{\mathrm{T}},\!\ActionSet_{\mathrm{T}},\!\Dynamics_{\mathrm{T}},\!g)$, where $\StateSet_{\mathrm{S}}\!\neq\!\StateSet_{\mathrm{T}}$, $\ActionSet_{\mathrm{S}}\!\neq\!\ActionSet_{\mathrm{T}}$, or $\Dynamics_{\mathrm{S}}\!\neq\!\Dynamics_{\mathrm{T}}$. 
%
Our objective is to optimize a policy $\pi_\theta$ from a single demonstration to maximize task success in the deployment domain:
\begin{equation}
\pi_{\theta}^{*} = \argmax_{\pi_{\theta}} 
\mathbb{E}_{\tau \sim p(\cdot \mid \pi_{\theta}, \mathcal{M}_{\mathrm{T}})} 
\big[
    \mathrm{SR}(\tau, g) - \lambda \mathrm{D}(\tau, \Demo)
\big]
\label{eq:objective}
\end{equation}
where $\tau\!=\!\{o_1, a_1, \dots, o_N, a_N\}$ denotes a trajectory sampled from $p(\tau\!\mid\!\pi_{\theta}, \mathcal{M}_{\mathrm{T}})$, induced by executing $\pi_\theta$ in $\mathcal{M}_{\mathrm{T}}$.
Here, $\mathrm{SR}(\tau, g)$ measures task success, $\mathrm{D}(\tau, \Demo)$ measures the deviation between $\tau$ and $\Demo$, and $\lambda$ is a weighting factor.
%
Thus, the policy $\pi_\theta^{*}$ aims to maximize task success in the deployment domain while maintaining alignment with the demonstration.
%
Each timestep $t$ corresponds to a semantically coherent interval, where the observation $o_t$ reflects the causal effect of action $a_t$~\cite{he2024learning, huey2025imitation}.
We represent $\pi_\theta$ as executable control code such that $\tau$ corresponds to its execution trace during deployment~\cite{cap:cap, vlcap:demo2code}.

\newcommand{\predSet}{\Psi}
\newcommand{\pred}{\psi}
\newcommand{\groundedPred}{\underline{\psi}}

\newcommand{\objectSet}{\mathcal{O}}
\newcommand{\sourceOperator}{\omega^{src}}
\newcommand{\sourceOperatorSet}{\Omega^{src}}
\newcommand{\targetOperatorSet}{\Omega^{tgt}}



\section{Neurosymbolic Counterfactual Reasoning}\label{sec:method}
%
We formulate video-instructed robotic programming as a cross-domain adaptation problem, in which the agent must transform an instructional video demonstration recorded into a logically verified and deployable task specification for execution in deployment domains with diverse environmental or embodiment conditions.
Such domain gaps alter how the agent interacts with its environment, thereby inducing procedural discrepancies between the demonstrated and deployable behaviors.
Rather than directly imitating the demonstrated behavior, the agent must reason about whether and how the demonstrated procedure should be revised under structural variations, thereby adjusting its actions to preserve task-level consistency.

We address this challenge through neurosymbolic counterfactual reasoning (\ours), which bridges the domain gaps between demonstrations and target observations by enabling procedure adaptation.
$\ours$ translates a demonstration into a symbolic representation of preconditions, actions, and effects, thereby constructing a symbolic world model. Given a target observation, the framework identifies its corresponding counterfactual situation by verifying which preconditions in the procedure are satisfied or violated. Based on this, $\ours$ adapts the demonstrated procedure by adding or removing actions so that the resulting procedure aligns the counterfactual state with the desired preconditions and effects.
Specifically, $\ours$ integrates VLM-based procedural alternative generation with symbolic verification and operates in two phases:
(\romannumeral 1) \textit{symbolic world model construction} and (\romannumeral 2) \textit{neurosymbolic counterfactual adaptation}.
In phase (\romannumeral 1), the demonstration is abstracted into a compact symbolic state sequence, from which a symbolic world model is constructed.
In phase (\romannumeral 2), the framework contrasts this symbolic world model with  target-domain observations to derive counterfactual states. The symbolic tool identifies where the demonstrated procedure fails, the VLM proposes alternatives to repair these steps, and the symbolic tool verifies their causal validity.
Through this neurosymbolic loop, $\ours$ produces a causally coherent, deployment-grounded task specification that is compiled into an executable code policy.

\begin{figure}[t]
\begin{center}
\includegraphics[width=.99\linewidth]{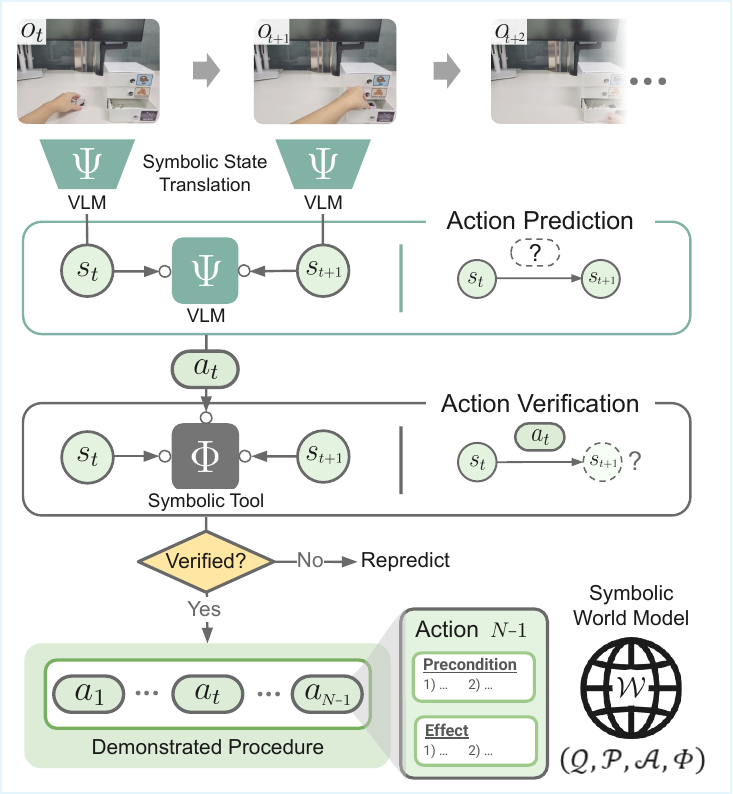}
\end{center}
\vspace{-15pt}
\caption{Symbolic world model construction}
\label{fig:method_1}
\vspace{-10pt}
\end{figure}

\subsection{Symbolic world model construction}\label{sec:method:1}
$\ours$ translates the demonstration into a symbolic world model that encodes the causal structure of the demonstrated behavior, ensuring its reproducibility within a symbolic state space.
As shown in Figure~\ref{fig:method_1}, rather than treating the demonstration as a raw sequence of observations, the VLM abstracts it into symbolic transitions. Consecutive observations are parsed into scene graphs representing symbolic states, from which the VLM predicts a symbolic operator that specifies the preconditions and effects of the executed action.
The symbolic tool then verifies the consistency of these transitions and integrates them into a unified world model that is logically coherent and reconstructable with respect to the demonstrated behavior. This symbolic world model $\mathcal{W}$ serves as a plan verification model and is expressed in a STRIPS-style formalism~\cite{strips}, supporting forward execution and logical validation.
\begin{equation}\label{symbolicworldmodel}
    \mathcal{W} = (\mathcal{Q}, \mathcal{P}, \mathcal{A}, \Phi), \quad \Phi : (s, a) \mapsto s'
\end{equation}
Here, $\mathcal{Q}$ denotes the set of objects identified in the scene, $\mathcal{P}$ the set of predicates representing object affordances and spatial relations, and $\mathcal{A}$ the set of symbolic actions, each defined by preconditions and effects over $\mathcal{P}$.
$\Phi$ denotes the symbolic tool (e.g., VAL~\cite{val, pddl2}) responsible for forward simulation and consistency verification.
Given a current symbolic state $s \in 2^{\mathcal{P}}$ and an action $a \in \mathcal{A}$ whose preconditions are satisfied in $s$, $\Phi$ applies the effects of $a$ to produce the next symbolic state $s'$.

\vspace{-7pt}
\paragraph{Symbolic state translation.}
To obtain a symbolic state sequence from the demonstration, $\ours$ prompts the VLM $\Psi$ to generate grounded scene graphs for key frames~\cite{scenegraph1, scenegraph2}.
At each timestep $t$, $\Psi$ extracts object entities and spatial relations from the image observation $o_t$ and language description $\instr$, forming a symbolic state $s_t$:
\begin{equation}\label{state}
\begin{aligned}
    & \Psi(\{o_1, \dots, o_N\}; \instr) = \{s_1, \dots, s_N\}, \\
    & \mathcal{Q} = \bigcup_{t=1}^N \text{Object}(s_t), \quad  \mathcal{P} = \bigcup_{t=1}^N \text{Predicate}(s_t)
\end{aligned}
\end{equation}
where $\text{Object}(s_t)$ and $\text{Predicate}(s_t)$ denote the objects and predicates grounded in $s_t$, respectively.
The resulting sequence $\{s_1,\!\dots,\!s_N\}$ represents a dynamic scene graph as a sequence of symbolic states derived from the demonstration, grounded in the object set $\mathcal{Q}$ and predicate set $\mathcal{P}$.

\vspace{-7pt}
\paragraph{Symbolic dynamics reconstruction.}
Given the symbolic state sequence $\{s_1,\!\dots,\!s_N\}$, $\ours$ abduces a set of symbolic action operators $\mathcal{A}$ that capture the causal transitions between consecutive states.
For each state pair $(s_t, s_{t+1})$, the VLM $\Psi$ predicts an action operator $a_t\!=\!\Psi(s_t, s_{t+1})$ whose effects correspond to the state difference $s_{t+1}\!\setminus\!s_t$, and whose preconditions hold in $s_t$.
Each $a_t$ is represented as a tuple $(\textit{name}, \textit{pre}, \textit{eff})$ and appended to $\mathcal{A}$.
To verify that $\mathcal{A}$ is consistent with the symbolic state sequence, the symbolic tool $\Phi$ performs forward simulation by applying each $a_t$ to $s_t$ and ensuring that the resulting symbolic state satisfies $s_{t+1}$ at every step.
\begin{equation}\label{dynamics}
    \big( \forall t \in [1, N{-}1],\; \Phi(s_t, a_t) \models s_{t+1} \big) \Rightarrow \text{Verified}(\mathcal{W})
\end{equation}
If this condition holds over the entire trajectory, $\mathcal{W}$ is constructed and verified, and the predicted action sequence is adopted as demonstrated procedure $\pi=\{a_1,\!\dots,\!a_{N-1}\}$.

\begin{figure}[t]
\begin{center}
\includegraphics[width=1.0\linewidth]{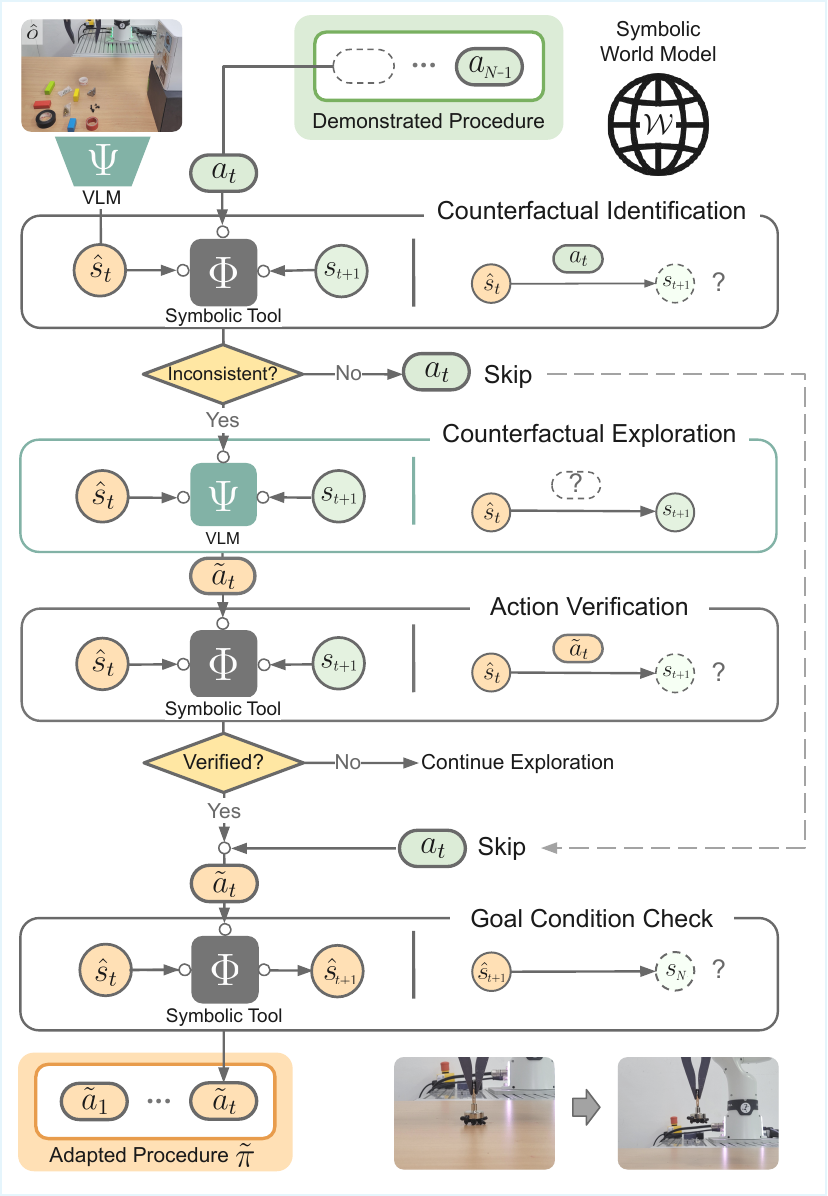}
\end{center}
\vspace{-15pt}
\caption{Neurosymbolic counterfactual adaptation}
\label{fig:method_2}
\vspace{-10pt}
\end{figure}

\subsection{Neurosymbolic counterfactual adaptation}\label{sec:method:2}
Based on the symbolic world model, $\ours$ performs counterfactual adaptation in a neurosymbolic manner, identifying causal inconsistencies induced by target-domain constraints and revising the demonstrated procedure to restore causal consistency.
As shown in Figure~\ref{fig:method_2}, given an observation from the deployment domain, the VLM generates a target symbolic state that serves as a counterfactual configuration by intervening on variables reflecting the target-domain conditions.
Using this counterfactual state, the symbolic tool performs forward simulation along the demonstrated procedure to identify actions whose preconditions are inconsistent with the counterfactual setting, thereby revealing cross-domain inconsistencies that hinder execution.
To resolve these conflicts, the VLM abduces alternative procedure steps through the insertion or removal of actions, while the symbolic tool iteratively verifies their logical consistency.
By iterating this exploration, the adapted procedure is refined into a deployment-grounded task specification, from which an executable code policy is synthesized.

\vspace{-7pt}
\paragraph{Counterfactuals identification.}
As the first step of neurosymbolic counterfactual adaptation, $\ours$ identifies causal inconsistencies in the demonstrated procedure through forward simulation under counterfactual conditions.
%
The VLM $\Psi$ generates a counterfactual state $\hat{s}_1$ that reflects the target observation, while the symbolic tool $\Phi$ simulates each action $a_t$ in the demonstrated procedure to estimate its outcome in the deployment domain.
\begin{equation}\label{rollout}
    \hat{s}_{t+1} = \Phi(\hat{s}_t, a_t), \quad t = 1, \dots, N{-}1
\end{equation}
An action is regarded as inconsistent when its preconditions are not satisfied in the current counterfactual state or when its effects fail to reproduce the expected predicates in the corresponding next state $s_{t+1}$.
%
\begin{equation}\label{conflict}
    \text{Inconsistent}(a_t) \Leftrightarrow \big(\textit{pre}(a_t)\!\nsubseteq\!\hat{s}_t \big) \vee \big(\textit{eff}(a_t)\!\nsubseteq\!s_{t+1} \big)
\end{equation}
Such inconsistencies in the procedure must be resolved to restore entire causal consistency in the deployment domain.

\vspace{-7pt}
\paragraph{Counterfactual exploration.}
To resolve these inconsistencies, $\ours$ performs counterfactual exploration within the symbolic state space, grounding the task procedure through additive and subtractive modifications.
For each inconsistent action identified, the VLM $\Psi$ proposes alternative actions whose effects restore the violated preconditions of the subsequent valid action $a_{t+1}$.
If no such alternative is applicable or the action becomes redundant to the task objective, it is removed from the procedure.
Let the action revised through counterfactual exploration be
\begin{equation}\label{interpolation}
    \tilde{a}_t =
    \begin{cases}
        \Psi(\hat{s}_t, a_t; s_{t+1} ), & \text{if } \text{Inconsistent}(a_t), \\
        a_t, & \text{otherwise.}
    \end{cases}
\end{equation}
The adapted procedure $\tilde{\pi}$, derived from Eq.~\eqref{interpolation}, restores procedural compatibility in deployment domain by satisfying
\begin{equation}\label{verification}
    \forall t \in [1, N{-}1], \quad \Phi(\hat{s}_t, \tilde{a}_t) = \hat{s}_{t+1}, \quad \text{until } \hat{s}_{t+1} \models s_N.
\end{equation}
Once $\tilde{\pi}$ reaches the goal condition specified by $s_N$, $\Psi$ translates $\tilde{\pi}$ into a code policy $\pi_\theta\!=\!\Psi(\tilde{\pi})$, ensuring that it remains grounded in the deployment domain.
Algorithm~\ref{alg} summarizes the overall process of $\ours$.

\begin{algorithm}[h]
\caption{$\ours$ Framework}\label{alg}
Demonstration $\Demo=(\{o_t\}_{t=1}^N,\instr)$, Target Observation $\hat{o}$ \\
VLM $\Psi$, Symbolic Tool $\Phi$, Symbolic Action Set $\mathcal{A}=\{\}$

\begin{algorithmic}[1]
\STATE \textcolor{blue}{\textit{/* Symbolic world model construction */}}
\STATE Get $\mathcal{Q}, \mathcal{P}$ and $\{s_t\}_{t=1}^N$ via Eq.~\eqref{state}
\FOR{$t=1, \dots, N{-}1$}
    \STATE Get grounded action $a_t=\Psi(s_t, s_{t+1})$
    \STATE Verify $a_t$ using $\Phi$; otherwise repredict $a_t$ \hfill \textcolor{gray}{\textit{cf.} Eq.~\eqref{dynamics}}
    \STATE $\mathcal{A} \leftarrow \mathcal{A}\cup\{a_t\}$
\ENDFOR
\STATE Get verified symbolic world model $\mathcal{W}=(\mathcal{Q}, \mathcal{P}, \mathcal{A}, \Phi)$
\STATE \textcolor{blue}{\textit{/* Neurosymbolic counterfactual adaptation */}}
\STATE Initialize adapted procedure $\tilde{\pi}=[ \ ]$, $t \leftarrow 0$
\REPEAT
    \STATE{$t \leftarrow t{+}1$; Get counterfactual state $\hat{s}_t = \Psi(s_t; \hat{o})$}
    \STATE{Do forward simulation using Eq.~\eqref{rollout}}
    \IF{$\text{Inconsistent}(a_t)$} 
        \STATE{Get interpolated action $\tilde{a}_t = \Psi(\hat{s}_t, a_t; s_{t+1})$}
    \ELSE
        \STATE{$\tilde{a}_t = a_t$}
    \ENDIF \hfill \textcolor{gray}{\textit{cf.} Eq.~\eqref{conflict} \& Eq.~\eqref{interpolation}}
    \STATE{Verify $\tilde{a}_t$ using $\Phi$ and get $\hat{s}_{t+1}$} via Eq.~\eqref{verification}
    \STATE{Append $\tilde{a}_t$ to $\tilde{\pi}$}
\UNTIL{$\hat{s}_{t+1} \models s_N$}
\STATE{Synthesize code policy $\pi_\theta = \Psi(\tilde{\pi})$}
\end{algorithmic}
\end{algorithm}

\section{Evaluation}\label{sec:evalu}
We evaluate $\ours$ through experiments designed to address the following questions:
\textbf{(Q1)} How does $\ours$ perform compared to existing baselines in cross-domain demo-to-code settings?
\textbf{(Q2)} How robust is $\ours$ to increasing domain gaps between the demonstration and deployment?
\textbf{(Q3)} How does $\ours$ respond as the complexity gap between the demonstration and deployment tasks grows?
\textbf{(Q4)} What is the contribution of each component of $\ours$?

\begin{table*}[th]
\begin{center}
\begin{small}
\begin{adjustbox}{width=0.98\linewidth}
\begin{tabular}{l | ccc | ccc | ccc}
    \toprule
    \multirow{2}{*}{Method}
    & \multicolumn{3}{c}{\textit{Low-Complexity}} & \multicolumn{3}{|c|}{\textit{Medium-Complexity}}  & \multicolumn{3}{c}{\textit{High-Complexity}}  \\
    \cmidrule(rl){2-4} \cmidrule(rl){5-7} \cmidrule(rl){8-10}
    & SR  & GC  & PD  & SR  & GC  & PD  & SR  & GC  & PD  \\
    \midrule
    \multicolumn{10}{l}{\textbf{Cross-domain factor}: \textit{Obstruction} and \textit{Object affordance}}  \\
    \midrule
    Demo2Code 
    & 26.67$\pm$5.76 & 55.00$\pm$4.24 & 5.00$\pm$3.42 
    & 25.00$\pm$5.64 & 51.11$\pm$4.37 & 10.67$\pm$5.11 
    & 22.50$\pm$6.69 & 61.25$\pm$4.38 & 6.94$\pm$2.20 \\
    \rowcolor{lightgray!30}
    GPT4V-Robotics
    & 71.67$\pm$5.87 & 82.50$\pm$3.91 & 29.15$\pm$5.34 
    & 41.67$\pm$6.42 & 68.89$\pm$4.11 & 43.18$\pm$5.09 
    & 20.00$\pm$6.41 & 57.50$\pm$4.49 & 35.94$\pm$10.68 \\
    \rowcolor{lightgray!30}
    Critic-V 
    & 45.00$\pm$6.48 & 65.83$\pm$4.52 & 10.37$\pm$4.58 
    & 35.00$\pm$6.21 & 64.44$\pm$4.18 & 24.97$\pm$4.95 
    & 15.00$\pm$5.72 & 58.75$\pm$3.96 & 45.83$\pm$9.50 \\
    \rowcolor{lightgray!30}
    MoReVQA
    & 53.33$\pm$6.49 & 71.67$\pm$4.35 & 32.29$\pm$5.17 
    & 43.33$\pm$6.45 & 75.00$\pm$3.23 & 25.17$\pm$6.07 
    & 27.50$\pm$7.15 & 73.75$\pm$3.22 & 40.91$\pm$8.10 \\
    Statler
    & 61.67$\pm$6.33 & 75.83$\pm$4.37 & 38.92$\pm$4.86 
    & 41.67$\pm$6.42 & 71.67$\pm$3.62 & 49.07$\pm$5.95 
    & 5.00$\pm$3.49 & 60.62$\pm$3.09 & 84.38$\pm$15.62 \\
    LLM-DM
    & 51.67$\pm$6.51 & 67.50$\pm$4.87 & 66.02$\pm$4.19
    & 20.00$\pm$5.21 & 47.78$\pm$4.37 & 77.13$\pm$3.21
    & 12.50$\pm$5.30 & 53.12$\pm$4.49 & 91.25$\pm$5.45 \\
    \rowcolor{lightgray!30}
    $\ours$
    & \textbf{86.67}$\pm$4.43 & \textbf{92.50}$\pm$2.61 & 1.92$\pm$1.35 
    & \textbf{75.00}$\pm$5.64 & \textbf{90.56}$\pm$2.25 & 10.57$\pm$2.22 
    & \textbf{60.00}$\pm$7.84 & \textbf{83.12}$\pm$3.94 & 12.50$\pm$2.91 \\
    \midrule
    \multicolumn{10}{l}{\textbf{Cross-domain factor}: \textit{Kinematic configuration} and \textit{Gripper type}}  \\
    \midrule
    Demo2Code 
    & 33.33$\pm$6.14 & 36.67$\pm$6.04 & 0.00$\pm$0.00 
    & 25.00$\pm$5.64 & 28.33$\pm$5.70 & 0.00$\pm$0.00 
    & 20.00$\pm$6.41 & 20.00$\pm$6.41 & 7.81$\pm$2.29 \\
    \rowcolor{lightgray!30}
    GPT4V-Robotics
    & 60.00$\pm$6.38 & 68.33$\pm$5.44 & 17.22$\pm$4.91 
    & 56.67$\pm$6.45 & 81.67$\pm$3.01 & 38.86$\pm$4.38 
    & 30.00$\pm$7.34 & 76.88$\pm$3.40 & 41.15$\pm$9.50 \\
    \rowcolor{lightgray!30}
    Critic-V 
    & 36.67$\pm$6.27 & 54.17$\pm$5.22 & 0.00$\pm$0.00 
    & 25.00$\pm$5.64 & 59.44$\pm$3.81 & 8.00$\pm$5.45 
    & 22.50$\pm$6.69 & 66.25$\pm$3.96 & 8.33$\pm$2.08 \\
    \rowcolor{lightgray!30}
    MoReVQA
    & 48.33$\pm$6.51 & 63.33$\pm$5.16 & 13.10$\pm$5.70 
    & 31.67$\pm$6.06 & 70.56$\pm$3.08 & 46.43$\pm$6.43 
    & 25.00$\pm$6.93 & 69.38$\pm$3.52 & 20.00$\pm$7.95 \\
    Statler
    & 68.33$\pm$6.06 & 74.17$\pm$5.25 & 29.43$\pm$5.14 
    & 50.00$\pm$6.51 & 70.56$\pm$4.35 & 53.35$\pm$4.31 
    & 42.50$\pm$7.92 & 74.38$\pm$4.24 & 50.00$\pm$5.60 \\
    LLM-DM
    & 53.33$\pm$6.49 & 69.17$\pm$4.77 & 58.33$\pm$5.32 
    & 35.00$\pm$6.21 & 62.22$\pm$4.15 & 81.69$\pm$3.70 
    & 27.50$\pm$7.15 & 67.50$\pm$4.12 & 76.14$\pm$9.79 \\
    \rowcolor{lightgray!30}
    $\ours$
    & \textbf{93.33}$\pm$3.25 & \textbf{96.67}$\pm$1.62 & 0.00$\pm$0.00 
    & \textbf{78.33}$\pm$5.36 & \textbf{85.00}$\pm$4.15 & 5.11$\pm$2.47 
    & \textbf{52.50}$\pm$8.00 & \textbf{78.12}$\pm$4.22 & 10.12$\pm$3.18 \\
    \midrule
    \multicolumn{10}{l}{\textbf{Cross-domain factor}: \textit{Combination} (\textit{Obstruction}, \textit{Object affordance}, \textit{Kinematic configuration}, and \textit{Gripper type})}  \\
    \midrule
    Demo2Code 
    & 30.00$\pm$7.34 & 46.25$\pm$6.55 & 0.00$\pm$0.00 
    & 20.00$\pm$6.41 & 48.33$\pm$5.96 & 5.56$\pm$3.64 
    & 7.50$\pm$4.22 & 28.75$\pm$5.77 & 12.50$\pm$6.25 \\
    \rowcolor{lightgray!30}
    GPT4V-Robotics
    & 55.00$\pm$7.97 & 70.00$\pm$5.88 & 15.76$\pm$5.80
    & 35.00$\pm$7.64 & 69.17$\pm$4.53 & 32.38$\pm$7.25 
    & 15.00$\pm$5.72 & 63.12$\pm$3.80 & 36.46$\pm$13.35 \\
    \rowcolor{lightgray!30}
    Critic-V 
    & 40.00$\pm$7.84 & 60.00$\pm$5.99 & 5.00$\pm$3.42
    & 37.50$\pm$7.75 & 69.17$\pm$4.68 & 5.93$\pm$3.41 
    & 27.50$\pm$7.15 & 66.25$\pm$4.44 & 16.48$\pm$4.15 \\
    \rowcolor{lightgray!30}
    MoReVQA
    & 55.00$\pm$7.97 & 72.50$\pm$5.36 & 14.85$\pm$4.88 
    & 45.00$\pm$7.97 & 68.33$\pm$5.46 & 48.49$\pm$8.61 
    & 25.00$\pm$6.93 & 62.50$\pm$4.56 & 25.62$\pm$6.81 \\
    Statler
    & 67.50$\pm$7.50 & 81.25$\pm$4.63 & 30.12$\pm$5.19 
    & 52.50$\pm$8.00 & 82.50$\pm$3.15 & 64.42$\pm$4.22 
    & 32.50$\pm$7.50 & 73.12$\pm$4.04 & 61.06$\pm$5.26 \\
    LLM-DM
    & 32.50$\pm$7.50 & 56.25$\pm$5.71 & 82.05$\pm$3.80 
    & 22.50$\pm$6.69 & 60.00$\pm$4.65 & 80.56$\pm$2.65 
    & 10.00$\pm$4.80 & 46.88$\pm$4.92 & 81.25$\pm$3.61 \\
    \rowcolor{lightgray!30}
    $\ours$
    & \textbf{80.00}$\pm$6.41 & \textbf{88.75}$\pm$3.79 & 2.81$\pm$1.97 
    & \textbf{65.00}$\pm$7.64  & \textbf{83.33}$\pm$4.30  & 6.60$\pm$2.53 
    & \textbf{47.50}$\pm$8.00 & \textbf{75.00}$\pm$4.48 & 17.11$\pm$3.16 \\
    \bottomrule
\end{tabular}
\end{adjustbox}
\end{small}
\end{center}
\vspace{-10pt}
\caption{
Performance on cross-domain demo-to-code tasks using simulation-based demonstration and deployment
}
\vspace{-5pt}
\label{tab:main:sim}
\end{table*}

\subsection{Experiment setting}
\paragraph{Cross-domain settings.}
Our cross-domain settings are defined by domain factors that introduce perceptual and physical variations between the demonstration and deployment domains.
We consider five such factors that induce procedural differences in task execution:
(1) \textit{Obstruction} introduces interfering objects that require additional resolving steps.
(2) \textit{Object affordance} alters object states and inter-object relations, yielding new affordances and relational dependencies.
(3) \textit{Kinematic configuration} changes the robot’s joint structure, affecting its reachable workspace and motion constraints.
(4) \textit{Gripper type} modifies the end-effector design, altering feasible grasp actions and contact affordances.
(5) \textit{Combination} jointly applies multiple domain factors, ranging from environmental factors (i.e., (1)-(2)) to embodiment factors (i.e., (3)-(4)), to create diverse and complex cross-domain scenarios. 
For demonstrations, we use data collected directly from both simulated and real environments, including robot executions and human-performed instructional videos.
For deployment, simulated experiments are conducted in Genesis~\cite{engine:genesis}, a physics-based general-purpose robotics platform, while real-world evaluations are performed on a {7-DoF Franka Emika Research~3}.

%

\vspace{-7pt}
\paragraph{Benchmark tasks.}
The deployment setting comprises long-horizon manipulation scenarios (up to 116 API calls per scenario) focused on table organization and object rearrangement~\cite{emb:vima, emb:robogen}.
Each scenario is composed of primitive subtasks (e.g., \textit{pick\&place}, \textit{sweep}, \textit{rotate}, \textit{slide}), forming procedurally compositional manipulation sequences. This compositional structure allows systematic control over task complexity.
For evaluation, scenarios are categorized into 3 complexity levels—\textit{Low}, \textit{Medium}, and \textit{High}—according to the number, diversity, and dependency depth of subtasks. Higher levels correspond to longer action horizons and stronger inter-subtask dependencies.
%
A total of 440 scenarios are categorized into three complexity levels: 160 for \textit{Low}, 160 for \textit{Medium}, and 120 for \textit{High}.

\vspace{-7pt}
\paragraph{Baselines.}
We implement six state-of-the-art baselines for video-instructed robotic programming, grouped into three:
(1) \textit{VLM-based code policy synthesis}, which generates task specifications from demonstrations and synthesizes code policies from the generated specifications, represented by Demo2Code~\cite{vlcap:demo2code},
(2) \textit{VLM-based reasoning}, which produces target-domain task specifications through the reasoning capabilities of VLMs, including GPT4V-Robotics~\cite{vlmreason:gpt4vrobot}, Critic-V~\cite{vlmreason:criticv}, and MoReVQA~\cite{vlmreason:morevqa}, and
(3) \textit{World-model-based approaches}, which construct LLM-based or neurosymbolic world models that support the generation of target task specifications, including Statler~\cite{wm:statler} and LLM-DM~\cite{wm:llmpddl}.
The baselines in (2) and (3) are not designed for code policy synthesis; in our evaluation, we apply their adaptation mechanisms in task specification generation, from which code policies are synthesized.

\vspace{-7pt}
\paragraph{Evaluation metrics.}
To evaluate the objectives in Eq.~\eqref{eq:objective}, we use several metrics.
\textbf{Success Rate (SR)} is the percentage of tasks completed in full, with a task counted as successful only when all subtasks are achieved.
\textbf{Goal Condition (GC)} measures the proportion of success conditions achieved, reflecting the degree of subtask completion~\cite{emb:alfred}.
\textbf{Procedure Deviation (PD)} quantifies the alignment between the adapted and demonstrated procedures using a success-weighted, length-normalized edit distance over their subtask-achievement sequences~\cite{motiv:planstability, inan2023multimodal}.
%
%

\subsection{Main result}
To address \textbf{(Q1)}, we evaluate $\ours$ in a simulated cross-domain environment where domain gaps are systematically induced through controlled variations of our domain factors. This setting allows fine-grained and reproducible analysis of code policy performance under different environmental and embodiment changes.
As shown in Table~\ref{tab:main:sim}, we compare $\ours$ with six state-of-the-art baselines across environmental, embodiment, and combined cross-domain scenarios.
$\ours$ consistently outperforms all baselines in both SR and GC across every domain factor. 
On average, it achieves a 27.73\% higher SR and 15.16\% higher GC than GPT4V-Robotics, the strongest \textit{VLM-based reasoning} baseline. Similarly, $\ours$ improves SR by 24.77\% and GC by 13.20\% over Statler, the strongest \textit{world-model-based} baseline. These results demonstrate that $\ours$ provides substantially more robust cross-domain adaptation than existing methods.

The baselines in the \textit{VLM-based reasoning} category, including GPT4V-Robotics, Critic-V, and MoReVQA, struggle to maintain consistent task specifications under domain shifts. In particular, erroneous feedback generation in Critic-V and error propagation across multi-stage reasoning in MoReVQA lead to substantial performance degradation, causing 40.91\% and 32.95\% drops in SR, respectively.
Although Statler leverages symbolic state representations and supports VLM-guided simulative planning, its lack of symbolic tool integration and dependence on replanning from scratch result in substantial performance degradation, especially in high-complexity scenarios.
LLM-DM, while employing symbolic tools, depends on constructing complete domain knowledge from a single demonstration, which frequently produces invalid or illogical plans, resulting in a 41.82\% drop in SR.
%
Demo2Code, which lacks a dedicated adaptation mechanism, exhibits a 49.09\% drop in SR compared to $\ours$. In terms of PD, however, $\ours$ achieves a comparable score, with only a 1.73 average increase across domain factors. This is because Demo2Code tends to imitate the demonstration without accounting for the deployment domain, yielding procedurally similar but frequently unreliable code policies.


\begin{table}[h]
\begin{center}
\begin{small}
\begin{adjustbox}{width=0.98\linewidth}
\begin{tabular}{l | ccc}
    \toprule
    \multirow{2}{*}{Method}
    & \multicolumn{3}{c}{\textit{Medium-Complexity}}  \\
    \cmidrule(rl){2-4}
    & SR  & GC  & PD   \\
    \midrule
    \multicolumn{4}{l}{\textbf{Cross-domain factor}: \textit{Combination}} \\
    \midrule
    Demo2Code 
    & 0.00$\pm$0.00 
    & 25.00$\pm$3.57 
    & $-$ \\
    \rowcolor{lightgray!30}
    GPT4V-Robotics
    & 50.00$\pm$18.90 
    & 75.00$\pm$9.64 
    & 0.00$\pm$0.00 \\
    Statler
    & 50.00$\pm$18.90 
    & 67.86$\pm$15.45 
    & 42.86$\pm$24.74 \\
    \rowcolor{lightgray!30}
    $\ours$
    & \textbf{87.50}$\pm$12.50 
    & \textbf{98.21}$\pm$1.79 
    & 24.49$\pm$11.54 \\
    \bottomrule
\end{tabular}
\end{adjustbox}
\end{small}
\end{center}
\vspace{-10pt}
\caption{
Performance on cross-domain demo-to-code tasks using real-world demonstrations and deployment
}
\label{tab:main:real}
\vspace{-5pt}
\end{table}

Furthermore, we evaluate $\ours$ in real-world deployment using a physical robot, with results presented in Table~\ref{tab:main:real}.
Using demonstrations captured from human video recordings, we evaluate performance on the same task executed under an obstructive domain gap using a Franka Research 3 arm.
In the demonstration, a human organizes objects into two parallel drawers.
In the deployment domain, these drawers sit at a perpendicular angle, creating a mutual-interference constraint: one drawer cannot open unless the other is closed.
Because the objects are arranged in stacks requiring ordered handling, the agent must alternate drawer operations, correctly inferring and maintaining the interference constraint throughout the procedure.
%
$\ours$ effectively adapts to domain changes, achieving an 87.50\% higher SR than Demo2Code and a 37.50\% higher SR than both Statler and GPT4V-Robotics.
GPT4V-Robotics often overlooks the constraints imposed by the perpendicular orientation and the stacked-object setting, which require alternating drawer operations; instead, it attempts to open both drawers at once.
Statler maintains a world state but, lacking any explicit mechanism for enforcing this constraint, intermittently attempts to open both drawers at once throughout the procedure.
In contrast, $\ours$ explicitly encodes the interference as a symbolic precondition and uses symbolic tools to verify its satisfaction throughout the entire procedure.
This enables an algorithmic process for detecting and resolving incompatible procedural steps regardless of horizon length, supporting reliable real-world control.


\subsection{Analysis and ablation}


\vspace{-10pt}
\begin{figure}[h]
    \centering
    \begin{subfigure}[b]{0.99\linewidth}
        \centering
        \includegraphics[width=\linewidth]{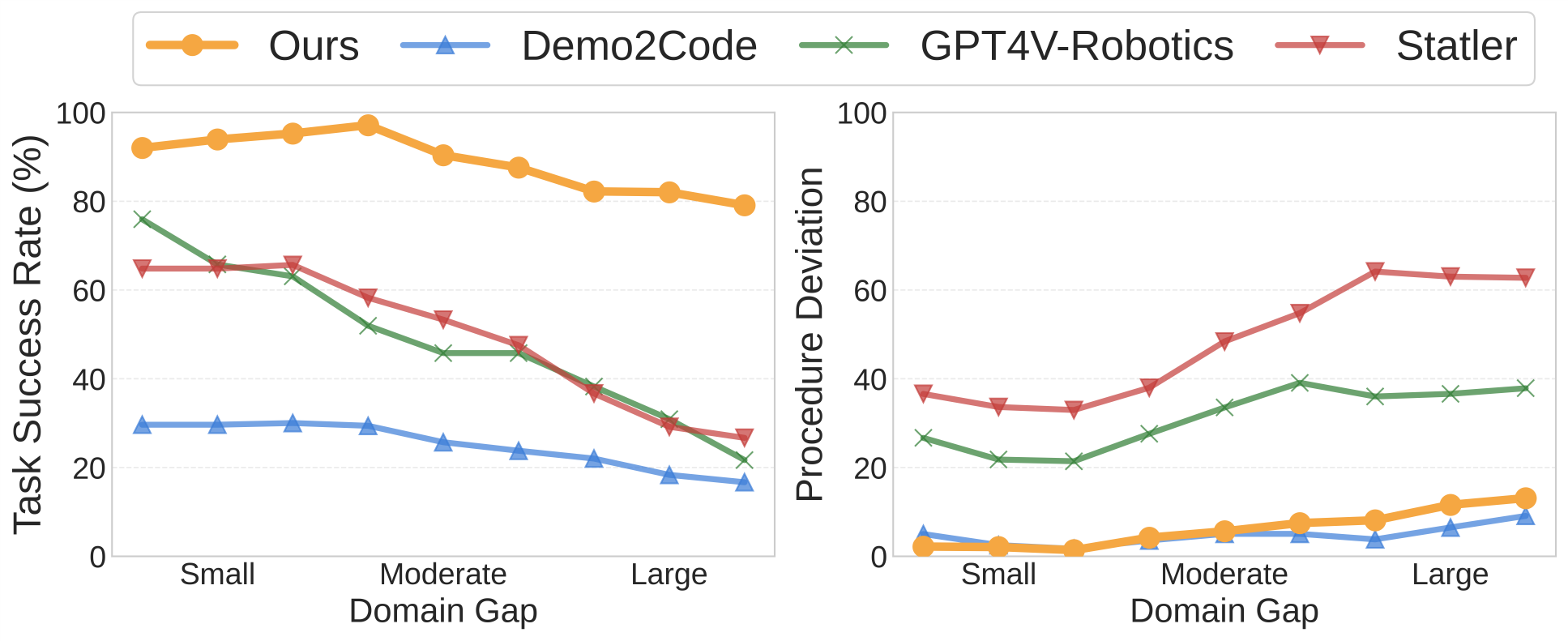}
        \caption{Comparison of SR, PD across different levels of domain gap}
        \label{fig:sub1}
    \end{subfigure}
    \begin{subfigure}[b]{0.99\linewidth}
        \centering
        \includegraphics[width=\linewidth]{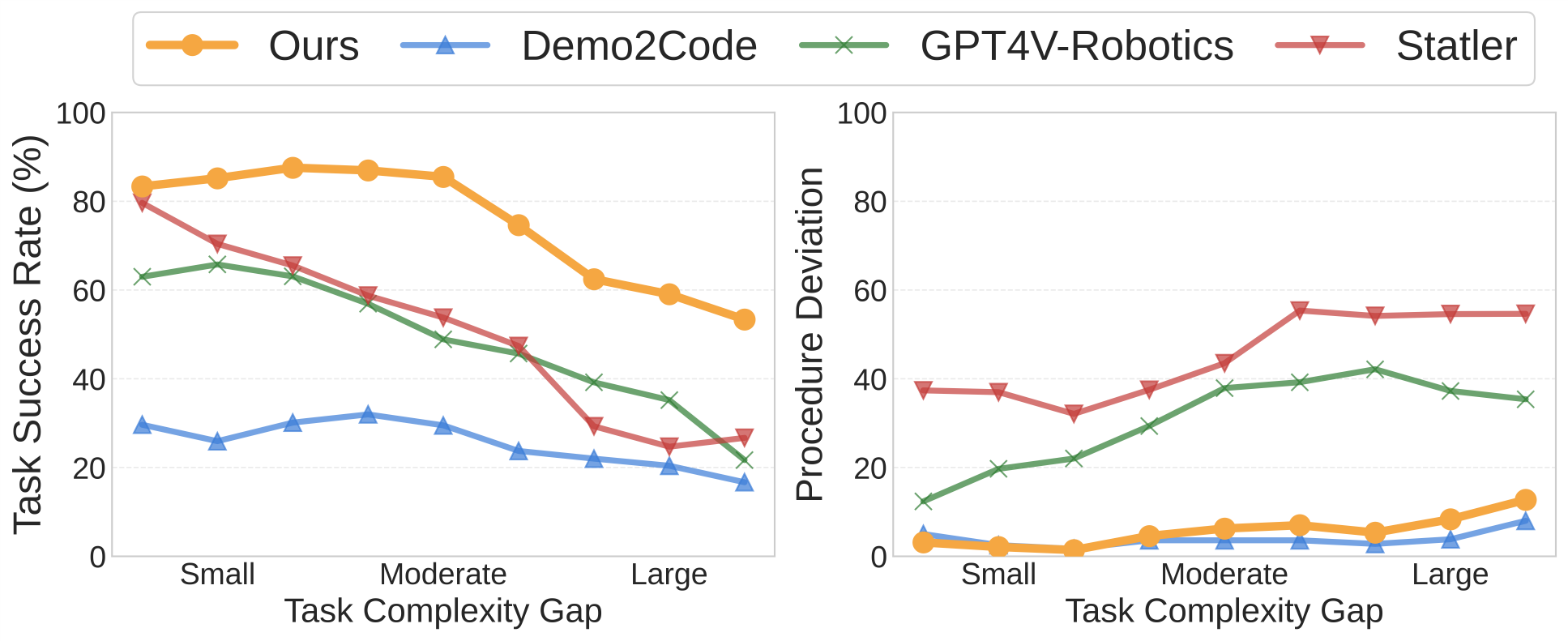}
        \caption{Comparison of SR, PD across different levels of task complexity gap}
        \label{fig:sub2}
    \end{subfigure}
    \vspace{-5pt}
    \caption{Analysis on (a) domain gap and (b) task complexity gap  
    }
    \label{fig:ab2}
    \vspace{-11pt}
\end{figure}

\vspace{-7pt}
\paragraph{Analysis on domain gap.}
To address \textbf{(Q2)}, we examine the robustness of $\ours$ as the domain gap between demonstration and deployment increases in Figure~\ref{fig:sub1}.
We amplify variations in environmental factors while keeping task complexity fixed, creating increasingly challenging cross-domain evaluation settings.
As the gap widens, the number of obstructing objects grows, requiring additional alternative actions to maintain procedural compatibility.
For all gap levels, $\ours$ explores and revises procedural steps more accurately than the baselines, achieving higher SR and lower PD. However, VLM-based baselines struggle to infer verified alternative action operators required to restore procedural compatibility, leading to a sharp performance drop from the Moderate to the High gap level.

\vspace{-7pt}
\paragraph{Analysis on task complexity gap.}
To address \textbf{(Q3)}, we evaluate the performance of $\ours$ under increasing task complexity gaps between the demonstration and deployment domains, as shown in Figure~\ref{fig:sub2}.
Task complexity is controlled by increasing the number of subtasks and the depth of their dependencies while fixing a domain factor such as \textit{Obstruction}.
When the complexity gap is moderate (e.g., up to 3 additional subtasks), $\ours$ effectively adapts the procedure. Yet, as the gap widens, the deployment-domain task scenario diverges from the demonstrated one, causing a pronounced performance drop as the demonstration no longer provides sufficient guidance and a fundamentally new demonstration becomes necessary.



\begin{table}[h]
\begin{center}
\begin{small}
\begin{adjustbox}{width=0.98\linewidth}
\begin{tabular}{l | ccc}
    \toprule
    \multirow{2}{*}{Method}
    & \multicolumn{3}{c}{\textit{Medium-Complexity}}  \\
    \cmidrule(rl){2-4}
    & SR  & GC  & PD   \\
    \midrule
    \multicolumn{4}{l}{\textbf{Cross-domain factor}: \textit{Combination}}  \\
    \midrule
    $\ours$
    & 68.42$\pm$7.64  
    & 84.21$\pm$4.49  
    & 6.60$\pm$2.53 \\
    $\ours$ w/o Eq.~\eqref{verification}
    & 50.00$\pm$8.22 
    & 78.07$\pm$4.03 
    & 14.65$\pm$3.20 \\
    $\ours$ w/o Eq.~\eqref{conflict}
    & 47.37$\pm$8.21 
    & 77.19$\pm$4.00 
    & 8.55$\pm$2.61 \\
    $\ours$ w/o Eq.~\eqref{conflict} \& Eq.~\eqref{verification}
    & 39.47$\pm$8.04 
    & 74.56$\pm$4.25 
    & 9.63$\pm$3.90 \\
    $\ours$ w/o Eq.~\eqref{dynamics}
    & 34.21$\pm$7.80 
    & 67.54$\pm$4.79 
    & 6.84$\pm$3.89 \\
    \bottomrule
\end{tabular}
\end{adjustbox}
\end{small}
\end{center}
\vspace{-12pt}
\caption{Ablation study of $\ours$}
\label{tab:q4}
\vspace{-13pt}
\end{table}

\vspace{-7pt}
\paragraph{Ablation study.}
To address \textbf{(Q4)}, we conduct an ablation study to assess the contribution of each component within $\ours$.
As shown in Table~\ref{tab:q4}, removing either the neurosymbolic counterfactual adaptation or the symbolic world model leads to a significant drop in SR and an increase in PD, underscoring the importance of these components for effective cross-domain demo-to-code performance.
Specifically, removing the verification of alternative actions (w/o Eq.~\eqref{verification}) results in an 18.42\% drop in SR, as the lack of verification allows the VLM to generate semantically plausible but logically inconsistent actions.
Removing the counterfactual identification (Eq.~\eqref{conflict}) prevents the framework from obtaining feedback from the symbolic tool on which procedural steps require revision, leading to a 21.05\% drop in SR.
When both components are removed (w/o Eq.~\eqref{verification} \& Eq.~\eqref{conflict}), SR decreases by 28.95\%. 
Removing the symbolic world model (w/o Eq.~\eqref{dynamics}) causes the most severe degradation, as the framework can no longer perform symbolic tool-based verification, resulting in the lowest SR of 34.21\%.


\begin{figure}[h]
\begin{center}
\includegraphics[width=0.90\linewidth]{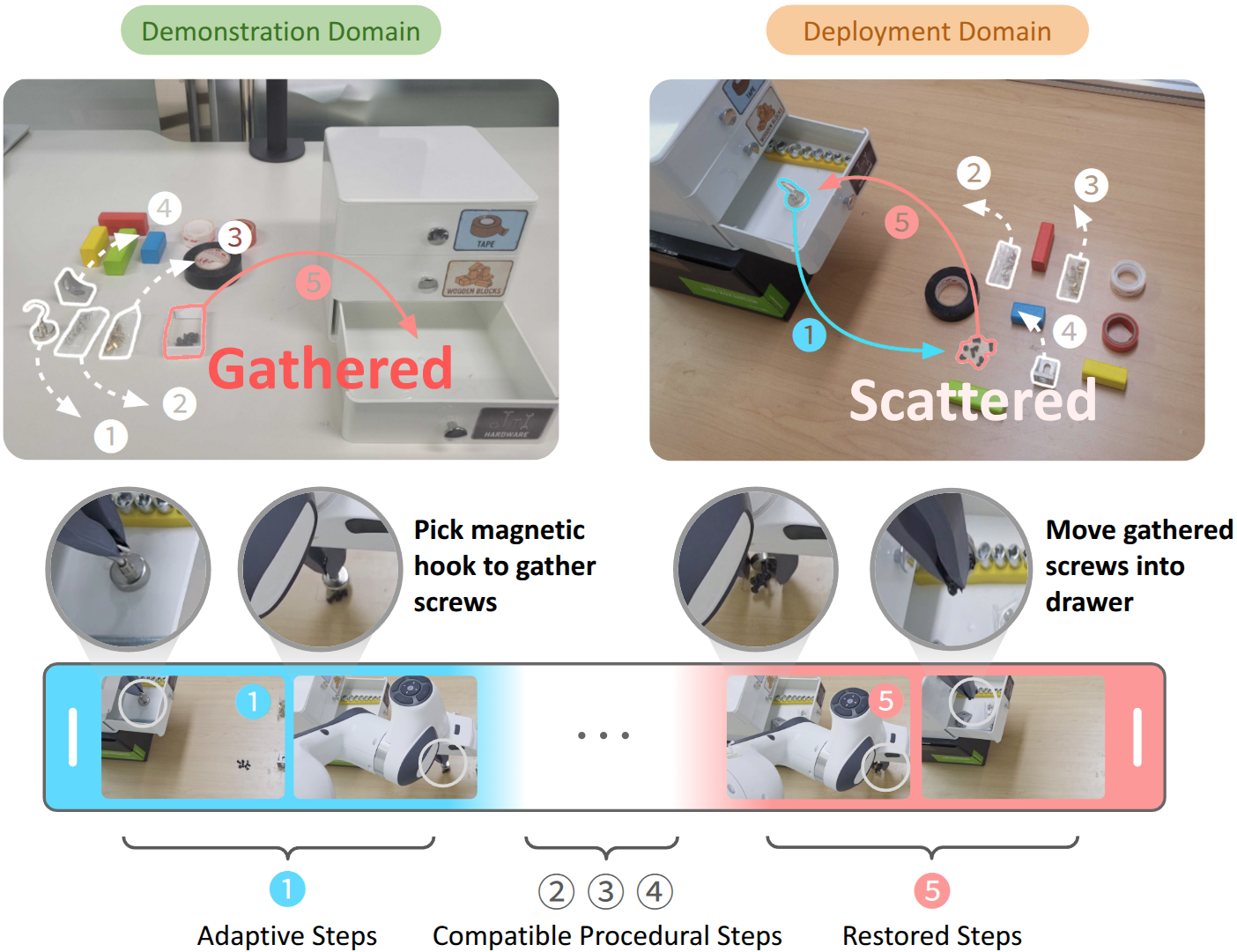}
\end{center}
\vspace{-15pt}
\caption{Visualization of a cross-domain demo-to-code task}
\label{fig:vis}
\vspace{-13pt}
\end{figure}

\vspace{-7pt}
\paragraph{Visualization of adaptation.}
To elaborate on the example in Figure~\ref{fig:overview}, we visualize the demonstration and the adapted procedure in the real-world deployment scenario.
As shown in Figure~\ref{fig:vis}, the demonstration domain features a human performing a drawer-organizing task with a magnetic hook and a box of screws on the table.
In the deployment domain, however, the human hand is replaced by a robotic gripper, the magnetic hook is already inside the drawer, and the screws are scattered across the surface.
First, $\ours$ detects that the magnetic hook already occupies the target position in the deployment domain and removes the redundant steps related to placing it. The VLM then repurposes the hook as an intermediate tool for aggregating the scattered screws, introducing actions for retrieving the hook and placing it over them.
These alternative steps, however, can be obstructed by objects that must also be placed in the drawer. $\ours$ resolves this potential failure by reordering the steps, performing the retrieval and aggregation steps before the organizing steps and producing a final procedure that restores compatibility with the deployment domain.

\section{Related work}
\label{sec:relat}
Foundation models provide commonsense knowledge and reasoning capabilities that enable generalized embodied control, leveraging pretrained language and vision knowledge to generate executable control code from instructions or demonstrations~\cite{cap:cap, cap:chatgptforrobotics, vlcap:demo2code, vlcap:robotic}.
Learning from demonstrations is common for embodied agents, but policies trained via behavioral cloning or inverse reinforcement learning struggle to generalize under perceptual and physical changes~\cite{LFD:survey, LFD:dagger, IFO, LFD:bcz}. Prior work has explored feature alignment, state-transition matching, and video-based imitation~\cite{VLFD:d3il, VLFD:dexmv, VLFD:oneshot, VLFD:visual}.
%
Neurosymbolic approaches combine neural adaptability with symbolic reasoning to improve correctness and interpretability in embodied planning, with recent methods using LLMs or VLMs to derive symbolic representations that are verified by symbolic solvers~\cite{nesyagent:nesyc, nesyagent:generalized, nesy:visualpredicator, nesy:worldcoder}, whereas our work integrates such neurosymbolic reasoning with counterfactual inference to revise demonstrated procedures for cross-domain adaptation.
Further discussion of related work is in Appendix A.

\section{Conclusion}\label{sec:concl}
In this work, we presented $\ours$, a neurosymbolic framework for cross-domain code synthesis in video-instructed robotic programming grounded in counterfactual reasoning. 
By constructing a verifiable symbolic world model and performing neurosymbolic counterfactual adaptation, $\ours$ converts video demonstrations into executable code policies that remain valid under perceptual and physical variations. 
Extensive experiments in simulation and real-world settings show that $\ours$ consistently outperforms existing baselines, maintaining higher task success and procedural consistency even as domain gaps and task complexity increase.
%
%
As shown in Figure~\ref{fig:ab2}, when the task complexity gap between the demonstration and deployment domains becomes large, the target task can no longer be solved via demonstration-based counterfactual reasoning alone.
Future work will extend this boundary through causal re-grounding, enabling $\ours$ to infer new causal relations and remain robust in novel contexts.

\section*{Acknowledgement}
This work was supported by Institute of Information \& communications Technology Planning \& Evaluation (IITP) grant funded by the Korea government (MSIT),
RS-2022-II220043, Adaptive Personality for Intelligent Agents,
RS-2022-II221045, Self-directed multi-modal Intelligence for solving unknown, open domain problems,
RS-2025-02218768, Accelerated Insight Reasoning via Continual Learning, and
RS-2025-25442569, AI Star Fellowship Support Program (Sungkyunkwan Univ.)
RS-2019-II190421, Artificial Intelligence Graduate School Program (Sungkyunkwan University)),
the National Research Foundation of Korea (NRF) grant funded by the Korea government (MSIT) (No. RS-2026-25474409),
IITP-ITRC (Information Technology Research Center) grant funded by the Korea government (MSIT) (IITP-2025-RS-2024-00437633, 10\%),
IITP-ICT Creative Consilience Program grant funded by the Korea government (MSIT) (IITP-2026-RS-2020-II201821), 
and by Samsung Electronics Co., Ltd.
{
    \small
    \bibliographystyle{ieeenat_fullname}
    \bibliography{main}
}

\onecolumn

\begin{center}
    {\Large \textbf{Cross-Domain Demo-to-Code via Neurosymbolic Counterfactual Reasoning}} \par
    \vspace{1.4em}
    {\Large \textit{Supplementary Material}}
\end{center}

\tableofcontents

\clearpage
\addtocontents{toc}{\protect\setcounter{tocdepth}{2}}
\appendix
\section{Further Related Works}\label{app:sec:rel}

\paragraph{Foundation models for embodied control.}
Foundation models have emerged as a promising alternative to conventional neural policies for embodied control, offering broad commonsense knowledge and reasoning capabilities that enable generalized task planning without extensive task-specific data~\cite{fdagent:saycan, fdagent:inner, fdagent:palme, fdagent:llmplanner, fdagent:emma, fdagent:capeam}. By leveraging pretrained language and vision representations, these models interpret human instructions and generate goal-directed behaviors for embodied agents. Recent studies have utilized LLMs and VLMs with code-writing capabilities to synthesize executable control code for embodied agents. LLM-based approaches map language instructions to code policies~\citep{cap:cap, cap:chatgptforrobotics, cap:instruct2act, cap:voxposer, cap:genchip, cap:promptbook, cap:gensim, cap:mccoder}, while VLM-based methods extend this to video demonstrations~\cite{vlcap:demo2code, vlcap:robotic, vlcap:seedo, vlcap:video2policy}. In this work, we enhance the VLM-based paradigm with robust cross-domain transfer to bridge mismatches between demonstrations and deployment.

\paragraph{Cross-domain adaptation from demonstrations.}
A well-established approach for training embodied agents is to learn from demonstrations, which enables policy acquisition without explicit rewards or manual supervision~\cite{LFD:survey, LFD:dagger, IFO, LFD:bcz}. However, policies learned through behavioral cloning or inverse reinforcement learning often struggle to generalize under significant domain shifts, such as variations in perceptual and physical factors, between expert demonstrations and the agent’s deployment~\cite{LFD, LFDRL, LFO, emb:conpe, emb:onis}. Prior works have explored adaptation strategies based on feature alignment, policy fine-tuning, and state-transition matching, as well as video-based imitation methods that align visual representations across domains (e.g., human-to-robot transfer)~\cite{VLFD:d3il, VLFD:dexmv, VLFD:oneshot, VLFD:visual}. Despite these, achieving robust generalization under perceptual and physical variations remains challenging, especially with limited demonstrations and procedurally complex tasks. In this work, we focus on domain gaps that induce procedural discrepancies and address them through counterfactual reasoning.

\paragraph{Neurosymbolic approaches for embodied agents.}
Neurosymbolic methods integrate the adaptability of neural networks with the formal reasoning capabilities of symbolic tools, improving the correctness and interpretability of task-level planning and reasoning. Recent studies have leveraged foundation models to produce symbolic formulations that are subsequently verified by symbolic tools~\cite{nesy:linc, nesy:logiclm, nesy:leveraging, nesy:coupling}. In embodied control, symbolic formalisms such as PDDL~\cite{pddl} and ASP~\cite{asp:solver} were employed, using LLMs to encode domain knowledge and translate task descriptions into problem instances~\cite{nesyagent:ai2thor, nesyagent:clmasp, nesyagent:generalized, nesyagent:llmp, nesyagent:llmrp, nesyagent:nesyc, nesyagent:nesypr, nesyagent:nesyro, nesy:casp, nesy:ivntr}. Furthermore, recent works have incorporated VLMs to enable perceptually grounded symbolic reasoning, leveraging the models’ vision–language pretraining to extract predicates, infer object relations, and construct symbolic world models that support embodied planning~\cite{nesy:psalm, nesy:visualpredicator, wm:pix2pred, nesy:worldcoder}. Our work integrates neurosymbolic reasoning with counterfactual inference, where the VLM proposes alternative actions while a symbolic tool validates and adjusts the resulting procedures to ensure reliable task execution across domains.

\section{Environment Settings}

\subsection{Genesis}
Genesis~\cite{engine:genesis} is a GPU-accelerated physics simulation platform designed for general-purpose robotic learning and evaluation. 
The platform provides a Python-based API that enables flexible implementation of custom environments with configurable scene layouts, object properties, and task specifications.
We leverage Genesis to construct the simulated environment for evaluating our models on the cross-domain demo-to-code tasks, as shown in Table 1 of the main paper.
Our evaluation framework encompasses two key dimensions: (1) cross-domain settings (rows of Table 1) that systematically vary environmental and embodiment factors, and (2) tasks with graduated complexity levels (columns of Table 1) ranging from simple operations to complex multi-step procedures. 
We provide detailed descriptions of each dimension below.

\subsubsection{Cross-domain settings}

To implement the cross-domain settings, we define five cross-domain factors grouped into three evaluation categories: (1) Obstruction and Object Affordance, which assess performance under environmental shifts; (2) Kinematic Configuration and Gripper Type, which assess performance under embodiment shifts; and (3) Combination, which evaluates robustness under both types of shifts simultaneously.
These cross-domain settings can be applied to each task, allowing to introduce controlled domain gaps between the demonstration and deployment domains. 
We describe each cross-domain setting in detail below.

\begin{figure*}[h]
\begin{center}
\includegraphics[width=.99\linewidth]{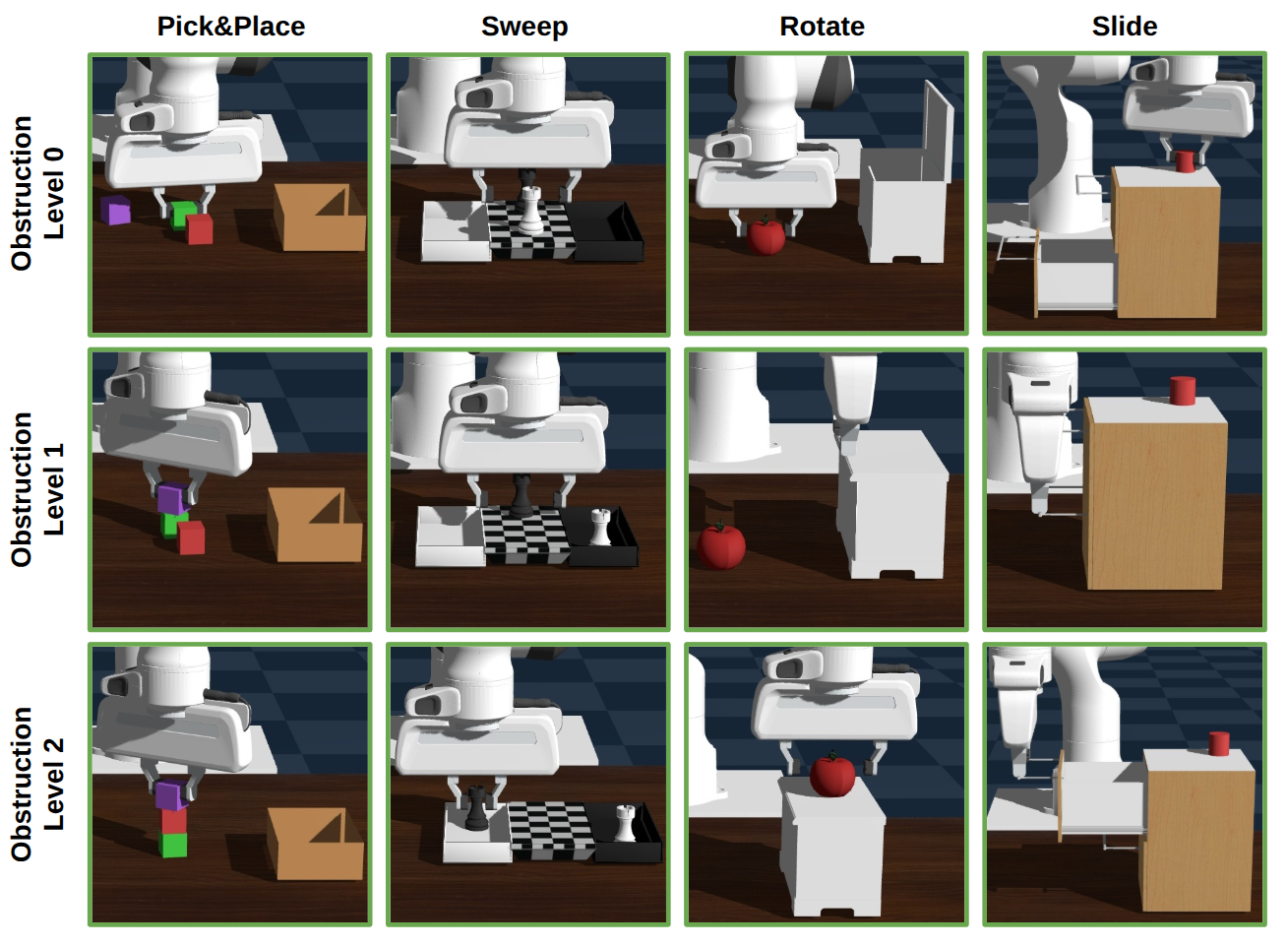}
\end{center}
\vspace{-15pt}
\caption{Example scene of Obstruction and Object affordance}
\label{app:fig:obstruction_level_description}
\end{figure*}

\paragraph{Obstruction and Object affordance.}
To implement this environmental factor, we introduce mechanisms to control obstruction and object-affordance levels for each subtask. The obstruction level ranges from 0 to 2, with higher levels indicating increased domain gap and task complexity.
As the obstruction level increases, target objects become more occluded, requiring the agent to resolve the obstruction before initiating the task. 
The scenes for the cross-domain settings for each subtask type are depicted in Figure~\ref{app:fig:obstruction_level_description}, with their explanations provided in Table~\ref{tab:obstruction_level_description}.

\begin{table*}[h]
\centering
\begin{tabular}{l|cc}
\toprule
\textbf{Task} & \textbf{Level 1} & \textbf{Level 2} \\
\midrule
Pick \& Place 
& A cube block stacked on preceding block.
& Multiple cube blocks stacked on preceding block. \\[3pt]

Sweep 
& A chess piece starts in the wrong box.
& Multiple chess pieces start in the wrong box. \\[3pt]

Rotate 
& Hinge lid starts closed. 
& Objects start stacked on top of the closed lid. \\[3pt]

Slide 
& Target drawer starts closed. 
& Target drawer starts closed, other obstructs from above. \\

\bottomrule
\end{tabular}
\caption{Description of task variations across obstruction levels}
\label{tab:obstruction_level_description}
\end{table*}

\begin{figure*}[h]
\begin{center}
\includegraphics[width=.5\linewidth]{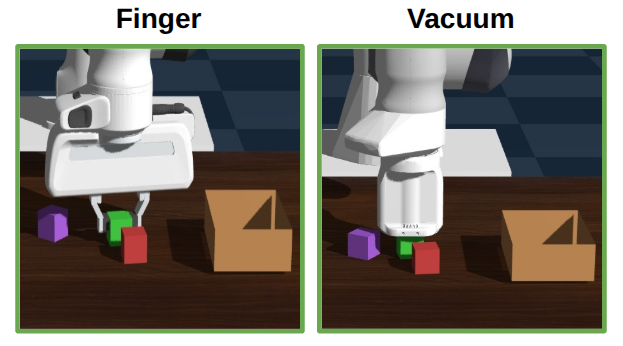}
\end{center}
\vspace{-15pt}
\caption{Example scene of Kinematic configuration and Gripper type}
\label{app:fig:kinematic}
\end{figure*}

\paragraph{Kinematic configuration and Gripper type.}
To implement this embodiment factor, we configure the robot with a 7-DoF vacuum suction gripper to compare it against a 9-DoF finger gripper. 
The finger gripper grasps objects by opening and closing its fingers, whereas the vacuum gripper secures objects by activating and deactivating its suction mechanism.
As each embodiment provides a distinct Action API set, the code policy must reorganize how actions are invoked and composed to conform to the target APIs. When adaptation is incomplete, this restructuring of API usage does not occur.
The example scenes for the cross-domain setting are depicted in Figure~\ref{app:fig:kinematic}.

\paragraph{Combination.}
The Combined setting implements all cross-domain factor variations simultaneously, including obstruction and object-affordance levels, kinematic configuration, and gripper type. 
This setup creates the most challenging evaluation condition by introducing domain gaps across both environmental and embodiment dimensions, requiring comprehensive adaptation from the demonstration to the deployment domain.

\subsubsection{Benchmark tasks}
\begin{figure*}[h]
\begin{center}
\includegraphics[width=.99\linewidth]{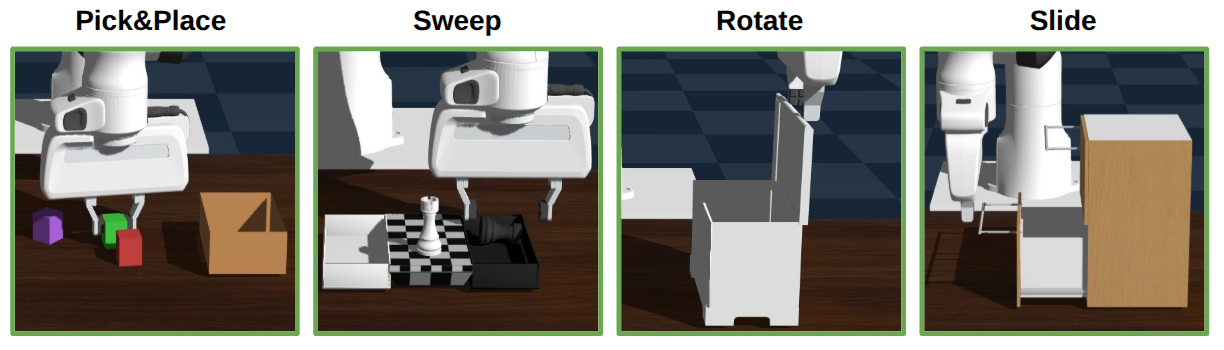}
\end{center}
\vspace{-15pt}
\caption{Example scenes for the subtasks}
\label{app:fig:subtask}
\end{figure*}

To assess the effectiveness of the generated code policy in the embodied domain, we design and implement four representative subtasks widely used in the area of robotic manipulation research~\cite{emb:vima, emb:robogen} in Genesis. 
These subtasks are composed to generate a single long-horizon task, with the difficulty of each subtask controlled by both domain factors and the complexity level, allowing for systematic evaluation.
The example scenes for each subtasks are depicted in Figure~\ref{app:fig:subtask}.

\begin{itemize}
    \item \textbf{Pick\&Place}. 
    The robot picks up cube blocks from the table and places them into a box. \\
    \item \textbf{Sweep}. 
    The robot sweeps chess pieces across the board, pushing each piece into its corresponding box. 
    The task succeeds only if all chess pieces are pushed inside their correct boxes. \\
    \item \textbf{Rotate}. 
    The robot picks up fruits, places them into a hinged container, and closes the container by rotating the lid around its axis. 
    The task is considered successful only if all fruits are placed inside the container and the lid is fully closed.\\ 
    \item \textbf{Slide}.
    The robot picks up cylinder blocks, places them into a drawer, and closes the drawer by sliding it shut. 
    The task is considered successful only if all cylinder blocks are placed inside the drawer and the drawer is fully closed.
\end{itemize}

\begin{figure*}[h]
\begin{center}
\includegraphics[width=.99\linewidth]{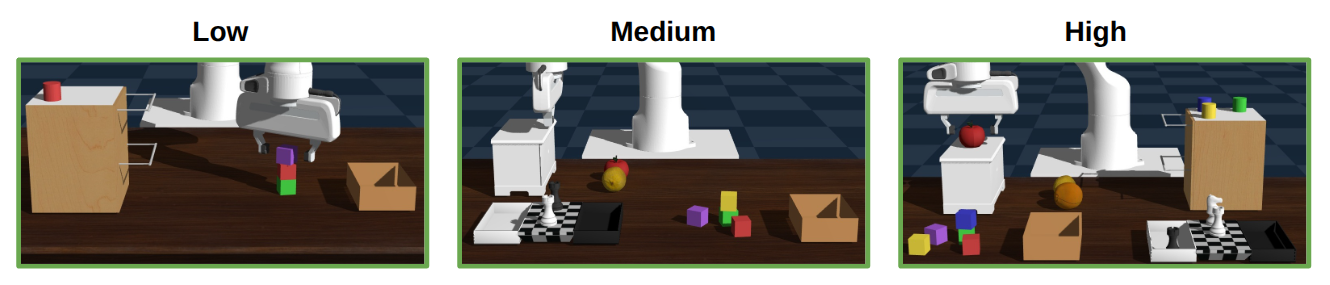}
\end{center}
\vspace{-15pt}
\caption{Example scenes for each task complexity level}
\label{app:fig:task_complexity}
\end{figure*}

\paragraph{Task complexity level.}
By controlling the individual complexity level of each subtask and the total number of subtasks used to compose a single task, we control the task complexity used for evaluation.
We divide the complexity level into three categories—low, medium, and high—where the low level consists of two subtasks, and the medium and high levels consist of three and four subtasks, respectively.
As complexity increases, not only does the number of subtasks grow, but the number of task-objects in each subtask also increases, making the task more complex and long-horizon.
Example scenes for each task complexity level are depicted in Figure~\ref{app:fig:task_complexity}.

\subsubsection{Evaluation}

We evaluate scenarios constructed by combining domain settings with task complexity levels. 
In total, 220 scenario configurations are generated and categorized into three complexity levels: 80 Low, 80 Medium, and 60 High. 
Each configuration is evaluated on scenes initialized with two different random seeds, yielding a total of 440 evaluation scenarios as depicted in Table~\ref{tab:task_distribution}.

\begin{table*}[h]
\centering
\begin{tabular}{l|ccc|c}
\toprule
\textbf{Domain Factors \textbackslash Task Complexity } & \textbf{Low} & \textbf{Medium} & \textbf{High} & \textbf{Total} \\ 
\midrule
Obstruction \& Object Affordance & 60 & 60 & 40 & 160 \\ 
Kinematic Config \& Gripper & 60 & 60 & 40 & 160 \\ 
Combination & 40 & 40 & 40 & 120 \\ 
\midrule
\textbf{Total Scenarios} & \textbf{160} & \textbf{160} & \textbf{120} & \textbf{440} \\ 
\bottomrule
\end{tabular}
\caption{Experimental scenarios for model evaluation}
\label{tab:task_distribution}
\end{table*}

\subsubsection{Primitive APIs}
To connect the generated code policy with robot control, we expose a set of primitive APIs that the VLM-generated code policy can invoke during execution. 
Specifically, we provide two primitive interfaces: a Perception API and an Action API.
The complete specifications of these APIs are detailed in Table~\ref{tab:perception_api} and Table~\ref{tab:action_api}.

\definecolor{darkgray}{rgb}{0.2,0.2,0.2}
\definecolor{lightgray}{rgb}{0.95,0.95,0.95}
\begin{table}[h]
\centering
\begin{tcolorbox}[
    colback=lightgray,
    colframe=darkgray,
    coltitle=white,
    title=Perception API,
    fonttitle=\bfseries,
    arc=1mm,
    width=0.9\linewidth,
    boxsep=2pt,
    left=3pt,
    right=3pt,
    top=3pt,
    bottom=3pt,
]
\centering
\setlength{\tabcolsep}{7pt}
\footnotesize
\begin{tabular}{p{4.2cm} p{9.5cm}}
\textbf{Primitive API} & \textbf{Description} \\ \midrule
\texttt{is\_obj\_visible(obj\_name)} & Returns a boolean indicating whether the object is present in the scene's object list. \\[16pt]
\texttt{get\_obj\_names()} & Returns a list of all object names currently present in the scene. \\[4pt]
\texttt{get\_obj\_pos(obj\_name)} & Returns the $(x, y, z)$ position of the corresponding object. \\[4pt]
\texttt{get\_obj\_bbox(obj\_name)} & Returns its axis-aligned bounding box \texttt{[min, max]} in world coordinates. \\[16pt]
\texttt{get\_obj\_size(obj\_name)} & Returns its size vector computed as the difference between the max and min corners of its bounding box. \\[16pt]
\texttt{gripper\_is\_open()} & Returns a Boolean indicating whether the gripper is currently open. \\[4pt]
\texttt{obj\_in\_gripper(obj\_name)} & Returns a Boolean indicating whether the object is currently within the gripper's grasp or suction region. \\[16pt]
\texttt{get\_empty\_floor\_xy(obj\_name)} & Returns a collision-free $(x, y)$ position on the floor where an object can be placed without overlapping existing objects. \\
\end{tabular}
\end{tcolorbox}
\vspace{-10pt}
\caption{Perception API primitives}
\label{tab:perception_api}
\end{table}

\centering
\begin{table*}[t]
\centering
\begin{tcolorbox}[
    colback=lightgray,
    colframe=darkgray,
    coltitle=white,
    title=Action API,
    fonttitle=\bfseries,
    arc=1mm,
    width=0.9\linewidth,
    boxsep=2pt,
    left=3pt,
    right=3pt,
    top=3pt,
    bottom=3pt,
]
\centering
\setlength{\tabcolsep}{7pt}
\footnotesize
\begin{tabular}{p{4.5cm} p{9.5cm}}
\textbf{Primitive API} & \textbf{Description} \\ \midrule
\texttt{move\_gripper\_to(obj\_name, depth)} & Moves the end-effector toward the object. \\[4pt]
\texttt{move\_to\_position(pos)} & Moves the end-effector to a target position. \\[4pt]
\texttt{move\_parallel(move\_dir, offset)} & Moves the end-effector parallel to the workspace plane in the specified direction by a given offset. \\[16pt]
\texttt{grasp\_handle(handle\_name)} & Grasps the handle when the end-effector is sufficiently close. \\[4pt]
\texttt{release\_handle()} & Releases the currently grasped handle and opens the gripper. \\[4pt]
\texttt{open\_gripper()} & Open the finger gripper. \\[4pt]
\texttt{close\_gripper()} & Close the finger gripper. \\[12pt]
\texttt{attach\_vacuum\_handle(handle\_name)} & Vacuum suction tool counterpart to \texttt{grasp\_handle}. \\[4pt]
\texttt{detach\_vacuum\_handle()} & Vacuum suction tool counterpart to \texttt{release\_handle}. \\[4pt]
\texttt{deactivate\_vacuum()} & Vacuum suction tool counterpart to \texttt{open\_gripper}. \\[4pt]
\texttt{activate\_vacuum()} & Vacuum suction tool counterpart to \texttt{close\_gripper}.
\end{tabular}
\end{tcolorbox}
\vspace{-10pt}
\caption{Action API primitives}
\label{tab:action_api}
\end{table*}
\raggedright


\subsection{Real-world}

\paragraph{Environment setup.}

We conducted the real-world experiments using a 7-DoF Franka Emika Research 3 robotic arm with a two-finger gripper on a tabletop workspace.
An Intel RealSense D435 RGB-D camera was mounted above the table to provide top-down RGB and depth observations.
The captured images were then processed by an object detection and segmentation module to extract the categories and bounding boxes of task-relevant objects.

\begin{figure*}[h]
\begin{center}
\includegraphics[width=.7\linewidth]{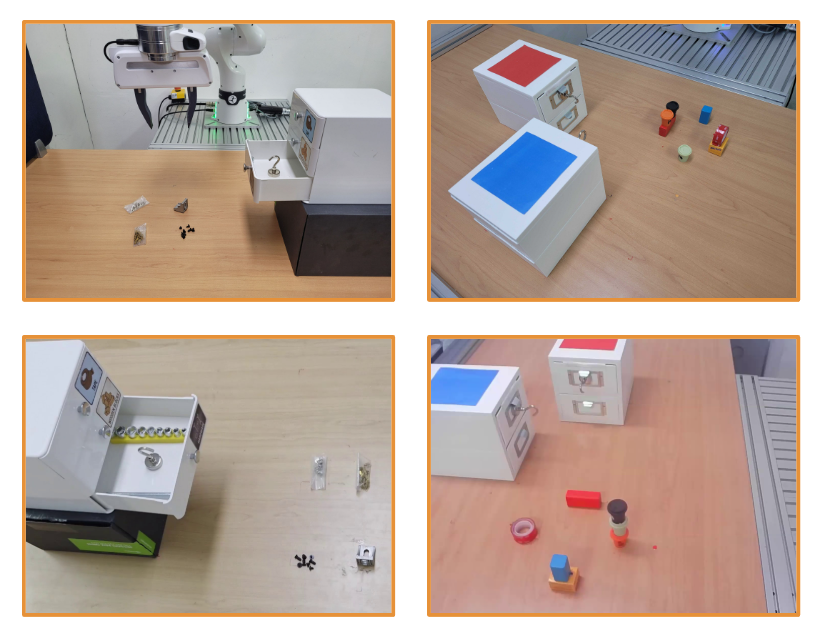}
\end{center}
\vspace{-15pt}
\caption{Example scenes from the real-world environment used in our experiments}
\label{app:real_world}
\end{figure*}

\paragraph{Object configuration.}

Across all experiments including the main scenario in Figure 1 of the main paper, we use an object pool consisting of three drawers, a magnetic hook, various types of screws, and other auxiliary objects for intermediate manipulations. 
The initial positions of all objects are randomized for each trial. Figure~\ref{app:real_world} shows a representative example of the environment used in the experiments.

\subsection{Metrics}

We employ three metrics that assess different aspects of task completion and procedure adherence.

\paragraph{Success Rate (SR).} 
The Success Rate measures the percentage of tasks that are completed in full. A task is counted as successful only when all subtasks are achieved:
$$
\text{SR} = \frac{1}{N} \sum_{i=1}^{N} \mathbbm{1}[\text{all subtasks completed in task } i]
$$
where $N$ is the total number of tasks and $\mathbbm{1}[\cdot]$ is the indicator function.

\paragraph{Goal Condition (GC).} 
The Goal Condition metric measures the proportion of success conditions achieved, reflecting the degree of subtask completion~\cite{emb:alfred}:
$$
\text{GC} = \frac{1}{N} \sum_{i=1}^{N} \frac{\text{number of achieved subtasks in task } i}{\text{total number of subtasks in task } i}
$$
This metric provides a more granular view of partial task completion compared to SR.

\paragraph{Procedure Deviation (PD).} 
The Procedure Deviation quantifies the alignment between the adapted procedure and the demonstrated procedure using a length-normalized edit distance over their subtask-achievement sequences of succeeded tasks~\cite{motiv:planstability, inan2023multimodal}.
$$
\text{PD} = \frac{1}{N_{\text{success}}} \sum_{i \in \mathcal{S}} \frac{\text{EditDistance}(S_i^{\text{adapt}}, S_i^{\text{demo}})}{\max(|S_i^{\text{adapt}}|, |S_i^{\text{demo}}|)}
$$
where $\mathcal{S}$ is the set of successfully completed tasks, $N_{\text{success}} = |\mathcal{S}|$ is the number of successful tasks, $S_i^{\text{adapt}}$ and $S_i^{\text{demo}}$ are the subtask-achievement sequences for the adapted and demonstrated procedures in task $i$, respectively, and $\text{EditDistance}(\cdot, \cdot)$~\cite{inan2023multimodal} computes the edit distance between two sequences.

\subsection{\textbf{$\ours$} Implementation}

In this section, we describe the implementation details of $\ours$, which is built upon two complementary reasoning engines: a vision language model (VLM) and a symbolic tool.
For the VLM component, we employ the general-purpose GPT-5 model and specialize its behavior via stage-specific prompting to satisfy the distinct functional requirements of each module.
For symbolic tool, we use the open-source PDDLGym~\footnote{https://github.com/tomsilver/pddlgym} library to handle PDDL parsing and state grounding. 
Because $\ours$ only requires forward state progression—rather than full planning—we implement a symbolic execution logic in Python that performs precondition verification and effect application.
We provide the pseudo-code for the symbolic tool in Algorithm~\ref{alg:symbolic}.

\begin{figure}[h]
\centering
\begin{minipage}{0.7\linewidth}
\begin{algorithm}[H]
\caption{Symbolic Tool $\Phi$}\label{alg:symbolic}
\begin{algorithmic}[1]
\STATE \textcolor{blue}{\textit{/* Forward execution with precondition check */}}
\STATE \textbf{function} \textsc{SymbolicExecute}($s, a$)
\begin{ALC@g}
\IF{$\text{pre}(a) \not\subseteq s$}
    \STATE \textbf{return} $\bot$, $\text{pre}(a) \setminus s$ \hfill \textcolor{gray}{\textit{Precondition Verification}}
\ENDIF
\STATE $s' \leftarrow (s \setminus \text{del}(a)) \cup \text{add}(a)$  \hfill \textcolor{gray}{\textit{Effect Application}}
\STATE \textbf{return} $s'$, $\emptyset$
\end{ALC@g}
\STATE
\STATE \textcolor{blue}{\textit{/* Verification process */}}
\STATE \textbf{function} \textsc{SymbolicVerify}($s, a$)
\begin{ALC@g}
\STATE $s', V \leftarrow $ \textsc{SymbolicExecute}($s, a$)
\IF{$V \neq \emptyset$}
    \STATE \textbf{return} \textsc{Fail}, $V$ \hfill \textcolor{gray}{\textit{Inconsistent}}
\ENDIF
\STATE \textbf{return} \textsc{Pass}, $s'$ \hfill \textcolor{gray}{\textit{Consistent}}
\end{ALC@g}
\end{algorithmic}
\end{algorithm}
\end{minipage}
\end{figure}

To use the symbolic representation, a predicate set that determines the scope of symbolic abstraction must be supplied.
We design a predicate set aimed at capturing generalizable relations common across embodied domains, including both physical relations and embodiment-specific predicates. 
Although richer or more domain-specific relations could be incorporated, we restrict ourselves to general predicates, as predicate invention lies outside the focus of this work. 
The predicate set used in main experiment is provided in Figure~\ref{fig:predicate_set}, note that this same predicate set is also shared among baselines that use symbolic representations.

\begin{figure*}[h]
\centering
\begin{tcolorbox}[
    colback=lightgray,
    colframe=darkgray,
    coltitle=white,
    title=Predicate Set,
    fonttitle=\bfseries,
    arc=1mm,
    width=0.7\linewidth,
    boxsep=2pt,
    left=3pt,
    right=3pt,
    top=3pt,
    bottom=3pt, 
    height=9cm, 
    valign=center, 
]
\centering
\setlength{\tabcolsep}{7pt}
\footnotesize
\begin{tabular}{p{3.4cm} p{4.6cm}}
\textbf{Predicate} & \textbf{Definition} \\ \midrule
\texttt{\# Physics-related} & \\ [4pt]
\texttt{(OverOf ?a ?b)} & \texttt{?a} is vertically above \texttt{?b} without contact. \\
\texttt{(OnTopOf ?a ?b)} & \texttt{?a} is resting on and supported by \texttt{?b}. \\
\texttt{(InsideOf ?a ?b)} & \texttt{?a} is contained within \texttt{?b}. \\[3pt]
\texttt{(Open ?x)} & Container \texttt{?x} is fully open. \\
\texttt{(Closed ?x)} & Container \texttt{?x} is fully closed. \\ [16pt]
\texttt{\# Embodiment-specific} & \\ [4pt]
\texttt{(FingerGripper)} & Robot uses a two-finger gripper. \\
\texttt{(VacuumSuction)} & Robot uses a vacuum suction tool. \\[3pt]
\texttt{(GripperSurrounding ?x)} & Gripper encloses \texttt{?x} without closing. \\
\texttt{(GripperHolding ?x)} & Gripper is closed and holding \texttt{?x}. \\
\texttt{(GripperOpen)} & Gripper is open. \\
\texttt{(GripperClosed)} & Gripper is closed. \\[3pt]
\texttt{(VacuumAligned ?x)} & Vacuum is aligned with \texttt{?x} but inactive. \\
\texttt{(VacuumAttached ?x)} & Vacuum is active and attached to \texttt{?x}. \\
\texttt{(VacuumActive)} & Vacuum suction is on. \\
\texttt{(VacuumInactive)} & Vacuum suction is off. \\
\end{tabular}
\end{tcolorbox}
\vspace{-10pt}
\caption{Predicate set used for symbolic state representation}
\label{fig:predicate_set}
\end{figure*}

\subsubsection{Symbolic World Model Construction}

\begin{figure}[H]\centering
\begin{varwidth}{0.9\linewidth}
\adjustbox{padding=10mm 8mm 10mm 8mm, frame}{
\begin{minipage}{0.8\linewidth}
\centering
\begin{tikzpicture}[
    scale=0.75,
    node distance=6mm and 8mm,
    every node/.style={font=\small},
    box/.style={draw, minimum height=6mm, inner sep=4pt, align=center},
    arrow/.style={-{Latex[length=1.5mm]}, thick}
]
    \node[box, fill=green!20] (demo) at (-6.0, 0) {Demonstration};
    \node[box, right=of demo] (states) {Symbolic\\States};
    \node[box, right=of states] (predict) {Action\\Operator};
    \node[box, fill=green!20, right=of predict] (model) {Demonstrated\\Procedure};
    
    \node[box, below=6mm of predict] (verify1) {Verification};
    
    \draw[arrow] (demo) -- node[above=4mm] {Translate} (states);
    \draw[arrow] (states) -- node[above=4mm] {Reconstruct} (predict);
    \draw[arrow] (predict) -- node[above=4mm] {Yield} (model);
    
    \draw[arrow] ([xshift=-2mm]predict.south) -- node[left=1mm] {Simulate} ([xshift=-2mm]verify1.north);
    \draw[arrow] (states.south) |- (verify1);
    \draw[arrow] ([xshift=2mm]verify1.north) -- node[right=1mm] {Verify} ([xshift=2mm]predict.south);
    
\end{tikzpicture}
\end{minipage}
}
\end{varwidth}
\caption{$\ours$ symbolic world model construction high-level flow}
\end{figure}
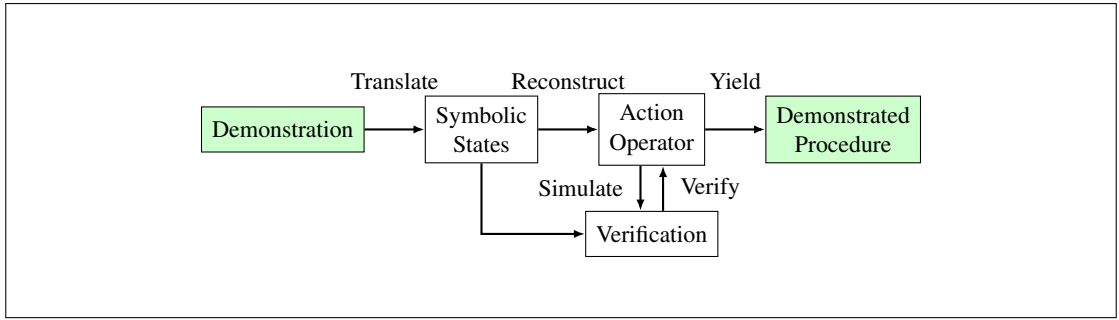

\begin{figure}[h]
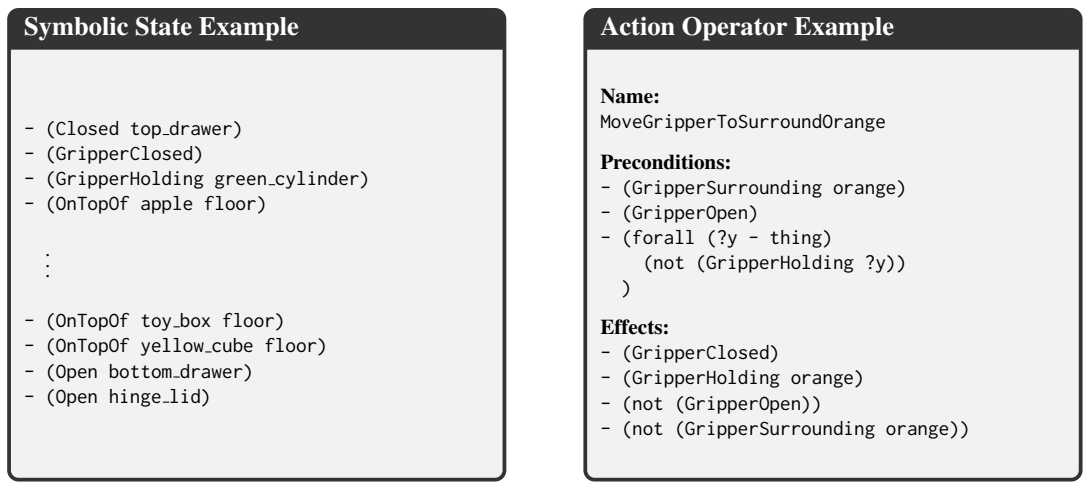

\centering
\begin{minipage}[t]{0.4\textwidth}
\begin{tcolorbox}[
    colback=lightgray,
    colframe=darkgray,
    coltitle=white,
    title=Symbolic State Example,
    fonttitle=\bfseries,
    arc=1mm,
    width=0.95\linewidth,
    boxsep=2pt,
    left=3pt,
    right=3pt,
    top=3pt,
    bottom=3pt,
    height=6.3cm, 
    valign=center, 
]
\footnotesize
\texttt{- (Closed top\_drawer)} \\
\texttt{- (GripperClosed)} \\
\texttt{- (GripperHolding green\_cylinder)} \\
\texttt{- (OnTopOf apple floor)} \\
\par
{\leftskip=1em
$\vdots$ \\
\par}
\texttt{- (OnTopOf toy\_box floor)} \\
\texttt{- (OnTopOf yellow\_cube floor)} \\
\texttt{- (Open bottom\_drawer)} \\
\texttt{- (Open hinge\_lid)}
\end{tcolorbox}
\end{minipage}
\hspace{0.5cm}
\begin{minipage}[t]{0.4\textwidth}
\begin{tcolorbox}[
    colback=lightgray,
    colframe=darkgray,
    coltitle=white,
    title=Action Operator Example,
    fonttitle=\bfseries,
    arc=1mm,
    width=0.95\linewidth,
    boxsep=2pt,
    left=3pt,
    right=3pt,
    top=3pt,
    bottom=3pt, 
    height=6.3cm, 
    valign=center, 
]
\footnotesize
\textbf{Name:} \\ 
\texttt{MoveGripperToSurroundOrange}

\vspace{2mm}
\textbf{Preconditions:} \\
\texttt{- (GripperSurrounding orange)} \\
\texttt{- (GripperOpen)} \\
\texttt{- (forall (?y - thing)} \\
\texttt{\hspace*{2em}(not (GripperHolding ?y))} \\
\texttt{\hspace*{1em})}

\vspace{2mm}
\textbf{Effects:} \\
\texttt{- (GripperClosed)} \\
\texttt{- (GripperHolding orange)} \\
\texttt{- (not (GripperOpen))} \\
\texttt{- (not (GripperSurrounding orange))}
\end{tcolorbox}
\end{minipage}

\caption{Example of symbolic state and action operator extracted from the demonstration}
\label{fig:symbolic_examples}
\end{figure}

\paragraph{VLM.}
In symbolic world model construction, VLM performs: (1) \textit{symbolic state translation}, where we process demonstration using a VLM with sequence of multi-view images and objects to extract grounded scene graphs representing symbolic states at each timestep, forming an ordered sequence of symbolic states from the first timestep to the final timestep that captures object entities and spatial relations; and (2) \textit{symbolic dynamics reconstruction}, where the VLM analyzes consecutive state pairs at each timestep to predict action operators. 
For each transition between timestep $t$ and $t+1$, the VLM is provided with the previous state, current state, and their state difference (additions and deletions of predicates).
The VLM then predicts action operators specifying: (i) action semantic description, (ii) preconditions required for execution, and (iii) effects produced after execution. 
Examples of the symbolic states and the action operators are provided in Figure~\ref{fig:symbolic_examples}.

\paragraph{Symbolic Tool.} 
In symbolic world model construction, symbolic tool performs: (1) \textit{action verification} through the following process: for each predicted action operator at timestep $t$, the tool (i) verifies that the current symbolic state fulfills all preconditions specified in the operator, (ii) applies the action's effects to the current state to produce the resulting next state, and (iii) validates that this resulting state matches the expected symbolic state at timestep $t+1$. 
If verification fails, the VLM is triggered to repredict the action operator until the entire sequence is verified.

\subsubsection{Neurosymbolic Counterfactual Adaptation}

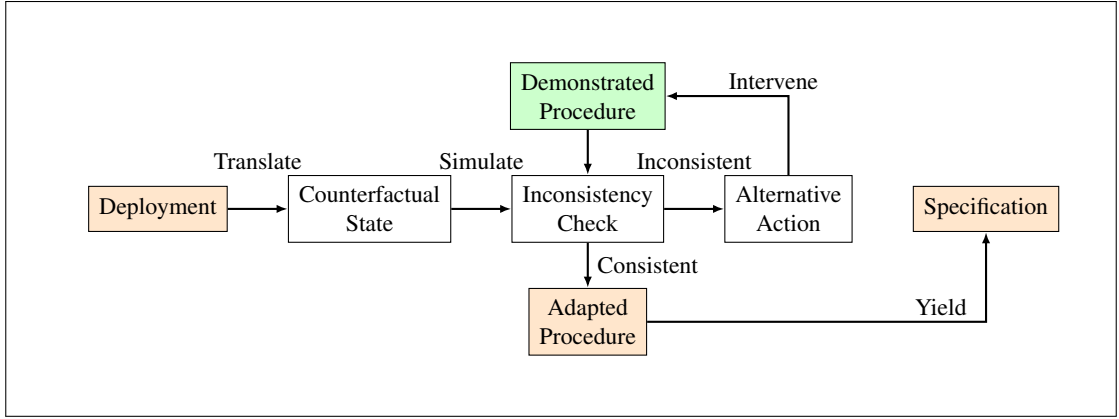
\begin{figure}[H]
\centering
\begin{varwidth}{0.9\linewidth}
\adjustbox{padding=10mm 8mm 10mm 8mm, frame}{
\begin{minipage}{0.8\linewidth}
\centering
\begin{tikzpicture}[
    scale=0.75,
    node distance=6mm and 8mm,
    every node/.style={font=\small},
    box/.style={draw, minimum height=6mm, inner sep=4pt, align=center},
    arrow/.style={-{Latex[length=1.5mm]}, thick}
]
    \node[box, fill=orange!20] (deploy) at (-6.0, -4.2) {Deployment};
    \node[box, right=of deploy] (cf) {Counterfactual\\State};
    \node[box, right=of cf] (identify) {Inconsistency\\Check};
    \node[box, right=of identify] (explore) {Alternative\\Action};
    \node[box, fill=green!20, above=of identify] (model) {Demonstrated\\Procedure};
    \node[box, fill=orange!20, below=of identify] (adapted) {Adapted\\Procedure};
    \node[box, fill=orange!20, right=of explore] (spec) {Specification};
    \draw[arrow] (deploy) -- node[above=4mm] {Translate} (cf);
    \draw[arrow] (cf) -- node[above=4mm] {Simulate} (identify);
    \draw[arrow] (identify) -- node[above=4mm] {Inconsistent} (explore);
    \draw[arrow] (model) -- (identify);
    \draw[arrow] (explore) |- node[shift={(-2mm, 2mm)}] {Intervene} (model);
    \draw[arrow] (identify) -- node[right=0mm] {Consistent} (adapted);
    \draw[arrow] (adapted.east) -| node[shift={(-6mm, 2mm)}] {Yield}(spec.south);
\end{tikzpicture}
\end{minipage}
}
\end{varwidth}
\caption{$\ours$ neurosymbolic counterfactual adaptation high-level flow}
\end{figure}

\paragraph{VLM.} 
In neurosymbolic counterfactual adaptation, VLM is used for: (1) \textit{counterfactual state translation}, where the VLM observes the deployment scene to generate a counterfactual initial state that reflects target-domain conditions while maintaining compatibility with the symbolic predicates from the world model; and (2) \textit{counterfactual exploration}, where for each action identified as inconsistent during symbolic forward simulation, the VLM proposes alternative action operators by analyzing the current counterfactual state, the incompatible original action, and the violated preconditions, generating interventions whose effects restore the violated preconditions or achieve equivalent outcomes under the deployment domain constraints.

\paragraph{Symbolic Tool.} 
In neurosymbolic counterfactual adaptation, the symbolic tool performs two functions: (1) \textit{counterfactual identification}, where it simulates the demonstrated procedure in the counterfactual setting by iteratively computing the next counterfactual state by applying each demonstrated action to the current counterfactual state, checking whether the action's preconditions are satisfied in the current state, flagging actions as inconsistent when some preconditions are violated; and (2) \textit{action verification}, where it validates VLM-proposed alternative action by verifying that its preconditions are satisfied in the counterfactual state.

\begin{figure*}[h]
\centering
\begin{minipage}[t]{0.4\textwidth}
\begin{tcolorbox}[
    colback=lightgray,
    colframe=darkgray,
    coltitle=white,
    title=Identification Example,
    fonttitle=\bfseries,
    arc=1mm,
    width=0.95\linewidth,
    boxsep=2pt,
    left=3pt,
    right=3pt,
    top=3pt,
    bottom=3pt,
    height=9.8cm, 
    valign=center, 
]
\footnotesize
\textbf{Current Counterfactual State:} \\ 
\texttt{- (Closed bottom\_drawer)}\\
\texttt{- (Closed hinge\_lid)}\\
\texttt{- ...}\\

\vspace{2mm}
\textbf{Incompatible Action:} \par
\vspace{1mm}
\texttt{OpenGripperToDropOrangeIntoHingeBody} \\[4pt]
\texttt{- Preconditions:} \\
\hspace*{1em}\texttt{- (not (Closed hinge\_lid))}\\
\hspace*{1em}\texttt{- ...} \\[4pt]
\texttt{- Effects:} \\
\hspace*{1em}\texttt{- (InsideOf orange hinge\_body)}\\
\hspace*{1em}\texttt{- ...} \\ 

\vspace{2mm}
\textbf{Violated Precondition:} \\
\texttt{- (not (Closed hinge\_lid))}\\
\end{tcolorbox}
\end{minipage}
\hspace{0.5cm}
\begin{minipage}[t]{0.4\textwidth}
\begin{tcolorbox}[
    colback=lightgray,
    colframe=darkgray,
    coltitle=white,
    title=Exploration Example,
    fonttitle=\bfseries,
    arc=1mm,
    width=0.95\linewidth,
    boxsep=2pt,
    left=3pt,
    right=3pt,
    top=3pt,
    bottom=3pt, 
    height=9.8cm, 
    valign=center, 
]
\footnotesize
\texttt{<<<<<<< SEARCH} \\ 
\texttt{OpenGripperToDropOrangeIntoHingeBody} \\[4pt]
\texttt{- Preconditions:} \\
\hspace*{1em}\texttt{- (not (Closed hinge\_lid))}\\
\hspace*{1em}\texttt{- ...} \\[4pt]
\texttt{- Effects:} \\
\hspace*{1em}\texttt{- (InsideOf orange hinge\_body)}\\
\hspace*{1em}\texttt{- ...} \par
{\leftskip=0em
$\vdots$ \\
\par}
\texttt{=======} \par
\texttt{MoveHeldOrangeOverFloor} \\[4pt]
\texttt{- Preconditions:} \\
\hspace*{1em}\texttt{- (GripperHolding orange)}\\
\hspace*{1em}\texttt{- ...} \\[4pt]
\texttt{- Effects:} \\
\hspace*{1em}\texttt{- (OverOf orange floor)}\\
\hspace*{1em}\texttt{- ...} \par
{\leftskip=0em
$\vdots$ \\
\par}
\texttt{>>>>>>> REPLACE} \\ 
\end{tcolorbox}
\end{minipage}

\caption{Example of counterfactual identification and exploration for the demonstration}
\label{fig:cf_examples}
\end{figure*}

The identification-exploration loop continues iteratively until either the adapted procedure successfully reaches the goal condition while maintaining causal consistency throughout, or the maximum number of iterations is reached.
An example of counterfactual identification and exploration is provided in Figure~\ref{fig:cf_examples}. 
Main hyperparameters of $\ours$ are listed in Table~\ref{app:hyper:nesycr}.

\begin{table*}[htbp]
\centering
\begin{tabular}{c c}
\toprule
\textbf{Hyperparameters} & \textbf{Value} \\
\midrule
\multirow{4}{*}{Max explorations} & 10 (low/medium), 20 (high) \\
 & {\small\textcolor{gray}{(Obstruction/Kinematic)}} \\[0.5em]
 & 15 (low/medium), 30 (high) \\
 & {\small\textcolor{gray}{(Combination)}} \\
\bottomrule
\end{tabular}
\caption{$\ours$ hyperparameters}
\label{app:hyper:nesycr}
\end{table*}

\clearpage
\subsection{Baselines}

\begin{figure*}[h]
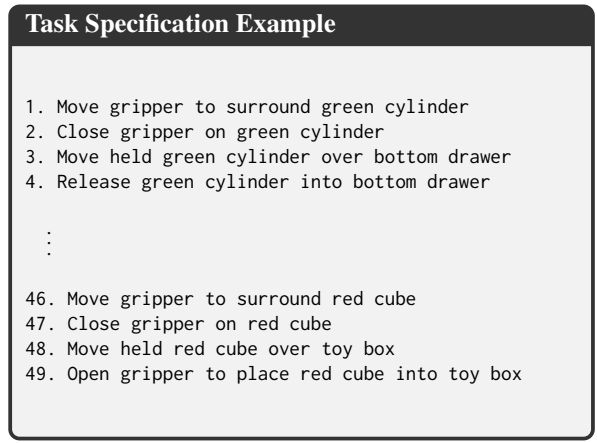

\centering
\begin{minipage}[t]{0.47\textwidth}
\begin{tcolorbox}[
    colback=lightgray,
    colframe=darkgray,
    coltitle=white,
    title=Task Specification Example,
    fonttitle=\bfseries,
    arc=1mm,
    width=0.95\linewidth,
    boxsep=2pt,
    left=3pt,
    right=3pt,
    top=3pt,
    bottom=3pt, 
    height=5.8cm, 
    valign=center, 
]
\footnotesize
\texttt{1. Move gripper to surround green cylinder} \\
\texttt{2. Close gripper on green cylinder} \\ 
\texttt{3. Move held green cylinder over bottom drawer} \\
\texttt{4. Release green cylinder into bottom drawer} \\ 
\par
{\leftskip=1em
$\vdots$ \\
\par}
\texttt{46. Move gripper to surround red cube} \\
\texttt{47. Close gripper on red cube} \\
\texttt{48. Move held red cube over toy box} \\
\texttt{49. Open gripper to place red cube into toy box}
\end{tcolorbox}
\end{minipage}
\caption{Example of task specification}  
\label{fig:task_spec}
\end{figure*}

All baselines receive a single demonstration as input—comprising a sequence of multiview images, instruction, and object set, supplemented with task context—and produce a target task specification in the form of high-level procedure.
Once the task specification is generated, the code policy synthesis process remains consistent across all baseline models.
An example of task specification is provided in Figure~\ref{fig:task_spec}.

Table~\ref{tab:baseline_comparison} summarizes the characteristics of each baseline method across five dimensions: 
\textbf{Natively Code} indicates whether the method generates code policies in its original work, or has been adapted for our evaluation.
\textbf{Explicit Adapt} denotes explicit mechanisms for adapting demonstrations to the deployment domain.
\textbf{Replanning} indicates generating compatible procedures by replanning from scratch in the deployment domain, rather than adapting demonstrations.
\textbf{World Model} refers to use of structured world representations.
\textbf{Symbolic Tool} denotes the use of external symbolic tool.

\begin{table*}[h]
\centering
\begin{tabular}{l|ccccc}
\toprule
\textbf{Baseline} & \textbf{Natively Code} & \textbf{Explicit Adapt} & \textbf{Replanning} & \textbf{World Model}  & \textbf{Symbolic Tool} \\
\midrule
Demo2Code & \cmark & \xmark & \xmark & \xmark & \xmark \\
GPT4V-Robotics & \cmark & \cmark & \cmark & \xmark & \xmark \\
Critic-V & \xmark & \cmark & \xmark & \xmark & \xmark \\
MoReVQA & \xmark & \cmark & \xmark & \xmark & \xmark \\
Statler & \xmark & \cmark & \cmark & \cmark & \xmark \\
LLM-DM & \xmark & \cmark & \cmark & \cmark & \cmark \\
\bottomrule
\end{tabular}
\caption{Comparison of baseline methods}
\label{tab:baseline_comparison}
\end{table*}

\begin{itemize}[leftmargin=*, itemsep=2pt, topsep=4pt]
    \item \textbf{VLM-based code policy synthesis.} uses VLMs to generate task specifications from visual demonstrations and synthesizes code policies, without incorporating an explicit adaptation mechanism.
    \begin{itemize}[leftmargin=*, itemsep=2pt, topsep=4pt]
        \item Demo2Code~\cite{vlcap:demo2code} generates task specifications by recursively summarizing the demonstration, with deployment domain information provided at intermediate steps.
    \end{itemize}
    
    \item \textbf{VLM-based reasoning.} utilize the reasoning capabilities of VLMs to perform adaptation, generating task specifications tailored to the target domain.
    \begin{itemize}[leftmargin=*, itemsep=2pt, topsep=4pt]
        \item GPT4V-Robotics~\cite{vlmreason:gpt4vrobot} generates task information from demonstrations and uses a VLM to generate a grounded target specifications from visually observing the target scene.
        \item Critic-V~\cite{vlmreason:criticv} uses a VLM-based critic to iteratively refine the initial task specification through generated critiques, ensuring compatibility with the deployment domain.
        \item MoReVQA~\cite{vlmreason:morevqa} follows a multi-stage modular reasoning process.
        It generates task information from demonstrations and uses VQA-style VLM querying to generate grounded target specifications.
    \end{itemize}
    
    \item \textbf{World-model-based approaches.} leverage LLM-based or neurosymbolic world models to support target task specification generation through high-level replanning mechanisms.
    \begin{itemize}[leftmargin=*, itemsep=2pt, topsep=4pt]
        \item Statler~\cite{wm:statler} equips LLMs with explicit world state representations that serve as memory throughout the replanning process, facilitating consistent reasoning over extended time horizons.
        \item LLM-DM~\cite{wm:llmpddl} constructs explicit PDDL world models from demonstrations using LLMs, then employs domain-independent symbolic planners to generate target task specifications by searching for high-level plans in the problem file for deployment domain.
    \end{itemize}
\end{itemize}

\paragraph{Demo2Code.} This method generates a code policy from demonstrations by recursively summarizing the demonstration through extended chain-of-thought to yield a task specification, which acts as a seed for generating the code policy.
While the original work does not assume a cross-domain setting, we modify the implementation by injecting deployment domain information in the final stage of summarization.

\begin{figure*}[h]
\centering
\begin{varwidth}{0.9\linewidth}
\setlength{\fboxsep}{10mm}
\fbox{%
\begin{minipage}{0.8\linewidth}
\centering

\begin{tikzpicture}[
    scale=0.75,
    node distance=6mm and 8mm,
    every node/.style={font=\small},
    box/.style={draw, minimum height=6mm, inner sep=4pt, align=center},
    arrow/.style={-{Latex[length=1.5mm]}, thick}
]
    \node[box, fill=green!20] (demo) {Demonstration};
    \node[box, right=of demo] (obs) {Observations};
    \node[box, fill=orange!20, below=of demo] (deploy) {Deployment};
    \node[box, right=of obs] (action) {Actions};
    \node[box, fill=orange!20, right=of action] (spec) {Specification};

    \draw[arrow] (demo) -- node[above=4mm] {Summarize} (obs);
    \draw[arrow] (obs) -- node[above=4mm] {Summarize} (action);
    \draw[arrow] (deploy) -| (action); 
    \draw[arrow] (action) -- node[above=4mm] {Yield} (spec);
    \draw[arrow] (action) -- (spec);
\end{tikzpicture}
\end{minipage}
}
\end{varwidth}
\caption{Demo2Code high-level flow}
\end{figure*}
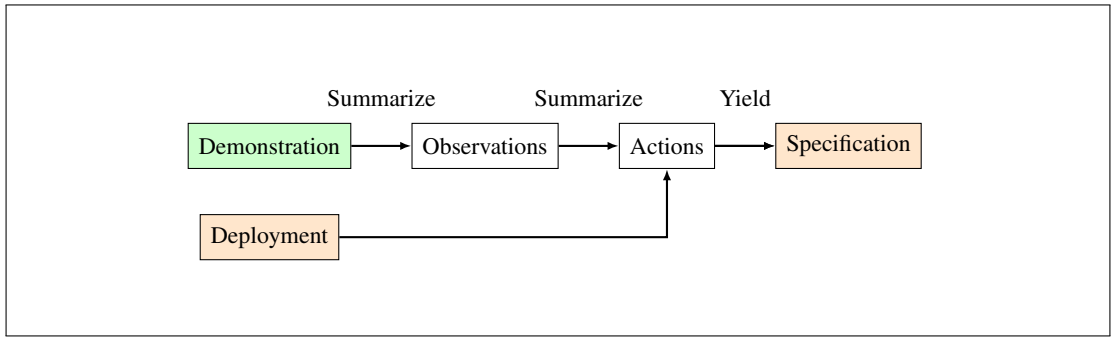

We implement Demo2Code referring to the official repository~\footnote{https://github.com/portal-cornell/demo2code}.
Implementation of adapted Demo2Code compose of: (1) \textit{observation prediction}, where we batch-process source demonstration timesteps using a VLM with multi-view images (top, front, back) to extract an ordered sequence of high-level observations; and (2) \textit{action prediction}, where the VLM predicts high-level actions from the observation sequence while being provided with deployment domain information to generate domain-adapted actions that serve as the task specification.

\paragraph{GPT4V-Robotics.} This method generates task specifications from visual demonstrations by first extracting domain-agnostic task descriptions from source demonstrations, then grounding them to deployment environments through visual scene understanding.
While the original work does not assume a cross-domain setting, we adapt the model by providing target scene information during the action planning stage.
Note that we retain the name GPT4V-Robotics from the original work, though we use GPT-5 as the base VLM for consistency.

\begin{figure*}[h]
\centering
\begin{varwidth}{0.9\linewidth}
\setlength{\fboxsep}{10mm}
\fbox{%
\begin{minipage}{0.8\linewidth}
\centering
\begin{tikzpicture}[
    scale=0.75,
    node distance=6mm and 8mm,
    every node/.style={font=\small},
    box/.style={draw, minimum height=6mm, inner sep=4pt, align=center},
    arrow/.style={-{Latex[length=1.5mm]}, thick}
]
    \node[box, fill=green!20] (demo) {Demonstration};
    \node[box, right=of demo] (domain) {Task\\Description};
    \node[box, fill=orange!20, below=of demo] (scene) {Deployment};
    \node[box, right=of domain] (actions) {Actions};
    \node[box, fill=orange!20, right=of actions] (spec) {Specification};
    
    \draw[arrow] (demo) -- node[above=4mm] {Extract} (domain);
    \draw[arrow] (domain) -- node[above=4mm] {Plan} (actions);
    \draw[arrow] (scene) -| (actions);
    \draw[arrow] (actions) -- node[above=4mm] {Yield} (spec);
\end{tikzpicture}
\end{minipage}
}
\end{varwidth}
\caption{GPT4V-Robotics high-level flow}
\end{figure*}
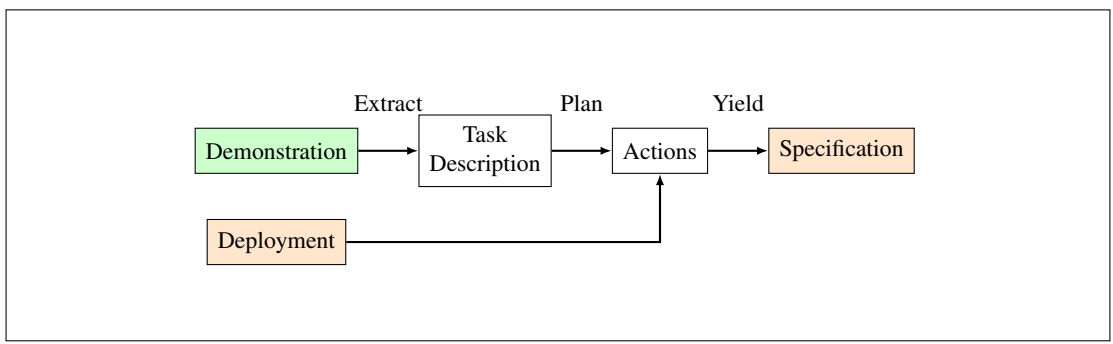

We implement GPT4V-Robotics referring to the official repository~\footnote{https://github.com/microsoft/GPT4Vision-Robot-Manipulation-Prompts}.
Implementation of adapted GPT4V-Robotics compose of: (1) \textit{task description generation}, where a VLM analyzes the source demonstration to extract high-level task understanding and domain knowledge; and (2) \textit{action planning}, where the VLM generates an ordered sequence of grounded actions by observing the deployment scene

\paragraph{Critic-V.} This method enhances VLM multimodal reasoning through iterative refinement with natural language critiques from visual observations. We adapt this framework for cross-domain demo-to-code by using it to iteratively refine action plans generated from source demonstrations until they align with the visual analysis of the deployment scene, and use the action plans to yield specifications for generating code policies.

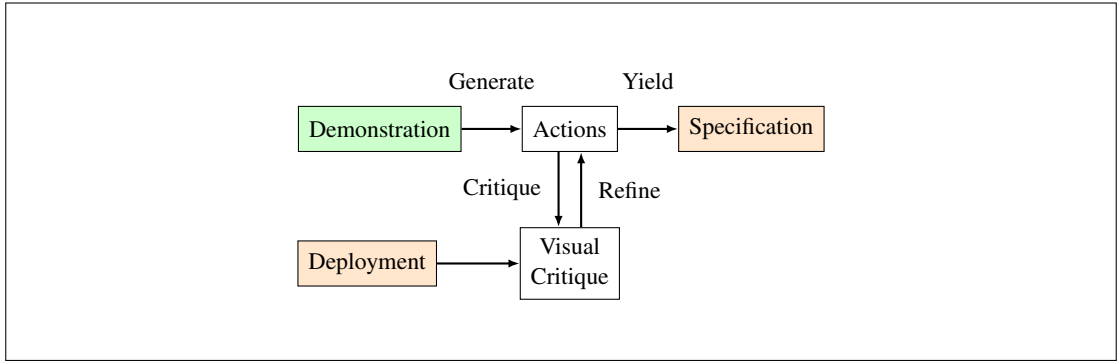
\begin{figure*}[h]
\centering
\begin{varwidth}{0.9\linewidth}
\adjustbox{padding=10mm 8mm 10mm 8mm, frame}{
\begin{minipage}{0.8\linewidth}
\centering
\begin{tikzpicture}[
    scale=0.75,
    node distance=6mm and 8mm,
    every node/.style={font=\small},
    box/.style={draw, minimum height=6mm, inner sep=4pt, align=center},
    arrow/.style={-{Latex[length=1.5mm]}, thick}
]
    \node[box, fill=green!20] (demo) {Demonstration};
    \node[box, right=of demo] (actions) {Actions};
    \node[box, fill=orange!20, right=of actions] (spec) {Specification};
    
    \node[box, below=10mm of actions] (feedback) {Visual\\Critique};
    \node[box, fill=orange!20, left=11mm of feedback] (deploy) {Deployment};
    
    \draw[arrow] (demo) -- node[above=4mm] {Generate} (actions);
    \draw[arrow] (actions)  -- node[above=4mm] {Yield} (spec);
    \draw[arrow] (deploy) -- (feedback);
    
    \draw[arrow] ([xshift=-2mm]actions.south) -- node[left=1mm] {Critique} ([xshift=-2mm]feedback.north);
    \draw[arrow] ([xshift=2mm]feedback.north) -- node[right=1mm] {Refine} ([xshift=2mm]actions.south);
\end{tikzpicture}
\end{minipage}
}
\end{varwidth}
\caption{Critic-V high-level flow}
\end{figure*}

We implement Critic-V referring to the official repository~\footnote{https://github.com/kyrieLei/Critic-V}. 
Implementation of the adapted Critic-V compose of: (1) \textit{initial action generation}, where a VLM generates an initial action sequence from source demonstrations; (2) \textit{critique generation}, where a VLM critic observes the deployment scene to identify incompatibilities in the actions and provides natural language feedback; and (3) \textit{action refinement}, where the VLM refines the actions based on the feedback.
Steps (2) and (3) repeat until the critic determines no issues exist or a maximum number of iterations is reached.

\begin{table*}[htbp]
\centering
\label{app:hyper:critic_v}
\begin{tabular}{c c}
\toprule
\textbf{Hyperparameters} & \textbf{Value} \\
\midrule
\multirow{4}{*}{Max critique refinement} & 10 (low/medium), 20 (high) \\
 & {\small\textcolor{gray}{(Obstruction/Kinematic)}} \\[0.5em]
 & 15 (low/medium), 30 (high) \\
 & {\small\textcolor{gray}{(Combination)}} \\
\bottomrule
\end{tabular}
\caption{Critic-V hyperparameters}
\end{table*}

\paragraph{MoReVQA.}
This method operated through a three-stage modular pipeline consisting of event parsing, grounding, and reasoning. We adapt it for cross-domain demo-to-code by using these stages to construct subgoals and query the deployment scene for achieve subgoal, ultimately generating a grounded specification that is used to produce the final code policy.

\begin{figure*}[h]
\centering
\begin{varwidth}{0.9\linewidth}
\setlength{\fboxsep}{10mm}
\fbox{%
\begin{minipage}{0.8\linewidth}
\centering
\begin{tikzpicture}[
    scale=0.75,
    node distance=4mm and 6mm,
    every node/.style={font=\small},
    box/.style={draw, minimum height=5mm, inner sep=4pt, align=center},
    arrow/.style={-{Latex[length=1.5mm]}, thick}
]
    \node[box, fill=green!20] (demo) {Demonstration};
    \node[box, right=5mm of demo] (m1) {M1\\Event Parsing};
    \node[box, right=5mm of m1] (m2) {M2\\Grounding};
    \node[box, right=5mm of m2] (m3) {M3\\Reasoning};

    \node[box, right=5mm of m3] (action) {Actions};
    \node[box, fill=orange!20, right=6mm of action] (spec) {Specification};

    \node[box, below=5mm of m1] (task) {Task\\Description};
    \node[box, fill=orange!20, below=7mm of m2] (deploy) {Deployment};
    
    \node[box, above=6mm of m2] (memory) {Shared\\Memory};
    \node[box, above=8mm of m3] (vqa) {VQA};
    
    \draw[arrow] (m3) -- node[above=2mm, xshift=4mm] {Generate} (action);
    \draw[arrow] (action) -- node[above=2mm] {Yield} (spec);
    
    \draw[arrow] (demo) -- (m1);
    \draw[arrow] (demo) |- node[shift={(8mm, 2mm)}] {Extract} (task);
    \draw[arrow] (task) -- (m2);
    
    \draw[arrow] (vqa) -- (m3);
    \draw[arrow] (deploy) -- (m2);
    
    \draw[arrow] (m1) -- (memory);
    \draw[arrow] (m2) -- (memory);

    \draw[arrow] (memory) -- (m3);
    \draw[arrow] (memory) -- (m2);
    \draw[arrow] (memory) -- (vqa);
\end{tikzpicture}
\end{minipage}
}
\end{varwidth}
\caption{MoReVQA high-level flow}
\end{figure*}
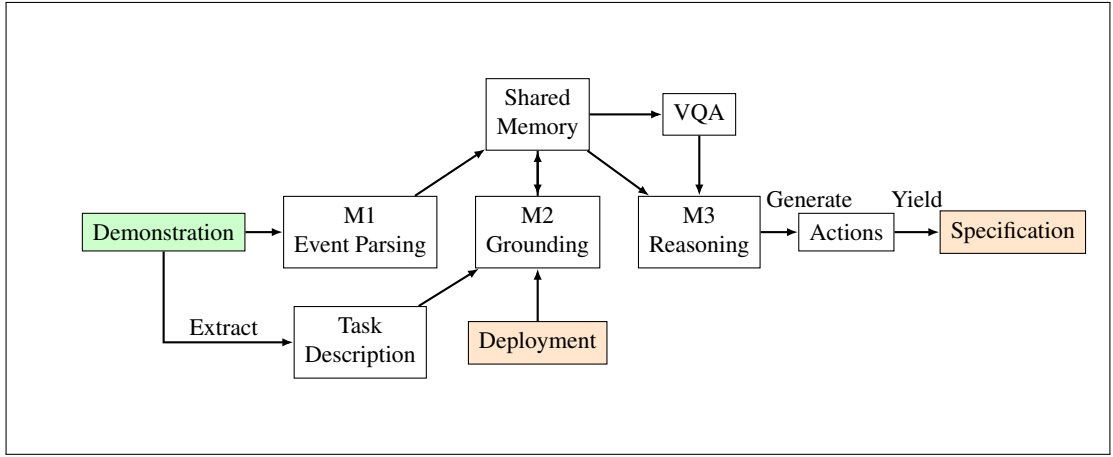

We implement MoReVQA based on the official supplementary material provided~\footnote{https://juhongm999.github.io/morevqa}.
The implementation of the adapted MoReVQA compose of: (1) \textit{M1 Event Parsing} processes the input instruction from the demonstration, converts it into a parsed event, and stores it in shared memory. (2) \textit{M2 Grounding} uses both the demonstration’s task description and the parsed event stored in shared memory to generate subgoals that are adapted to the deployment scene. (3) \textit{M3 Reasoning} combines the parsed event in memory with the VQA results derived from the deployment scene and generates the actions required to achieve each subgoal.

\paragraph{Statler.} This method ensures consistent state tracking across planning steps by maintaining an explicit world state representation as a memory. 
We adapt this framework for cross-domain demo-to-code scenarios by extracting task descriptions from demonstrations to re-generate action plans on the deployment domain, and use the action plans to yield specifications for generating code policies.

\begin{figure*}[h]
\centering
\begin{varwidth}{0.9\linewidth}
\adjustbox{padding=10mm 8mm 10mm 8mm, frame}{
\begin{minipage}{0.8\linewidth}
\centering
\begin{tikzpicture}[
scale=0.75,
node distance=6mm and 8mm,
every node/.style={font=\small},
box/.style={draw, minimum height=6mm, inner sep=4pt, align=center},
arrow/.style={-{Latex[length=1.5mm]}, thick}
]
\node[box, fill=green!20] (demo) {Demonstration};
\node[box, right=of demo] (domain) {Task\\Description};
\node[box, fill=orange!20, below=10mm of demo] (scene) {Deployment};
\node[box, right=11mm of scene] (state) {Predicted\\State};
\node[box, right=of state] (actions) {Actions\\(Incremental)};
\node[box, fill=orange!20, right=of actions] (spec) {Specification};

\draw[arrow] (demo) -- node[above=4mm] {Extract} (domain);
\draw[arrow] (scene) -- node[above=4mm] {Predict} (state);
\draw[arrow] (domain) -| (actions);
\draw[arrow] (state) -- node[above=4mm] {Plan} (actions);
\draw[arrow] (actions) -- node[above=4mm] {Yield} (spec);

\draw[arrow] ([xshift=-2mm]actions.south) .. controls +(down:8mm) and +(down:8mm) .. node[below=2mm] {Update} ([xshift=2mm]state.south);
\end{tikzpicture}
\end{minipage}
}
\end{varwidth}
\caption{Statler high-level flow}
\end{figure*}
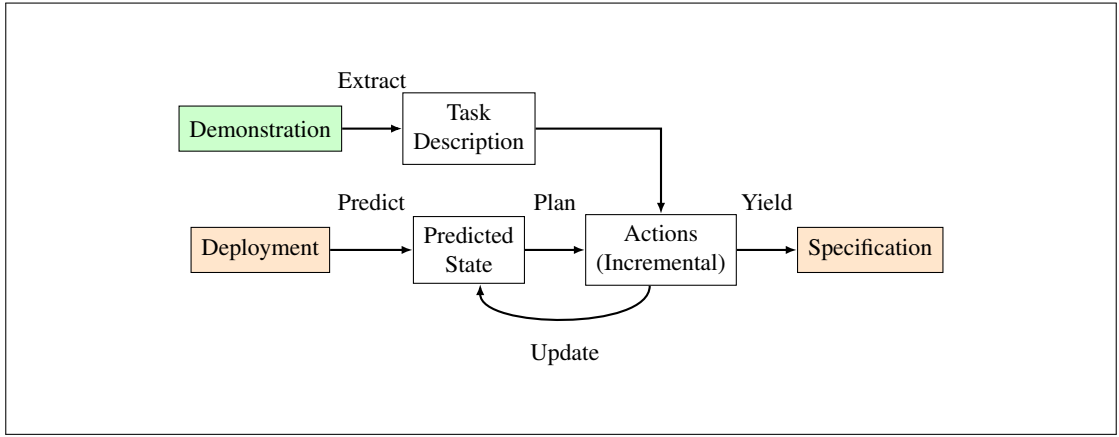

We implement Statler referring to the official repository~\footnote{https://github.com/ripl/statler}. 
Implementation of the adapted Statler compose of: (1) \textit{task description generation}, where a VLM analyzes the source demonstration to extract high-level task understanding and domain knowledge; (2) \textit{initial state prediction}, where the VLM observes the deployment scene to predict the initial state using the same predefined predicates as $\ours$; and (3) \textit{incremental action planning}, where at each step the VLM generates next several actions conditioned on the current state, task description, and previous actions, then updates the state representation accordingly. 
This state-action cycle repeats iteratively until the goal is reached or a maximum number of iterations is reached.

\begin{table*}[htbp]
\centering
\label{app:hyper:statler}
\begin{tabular}{c c}
\toprule
\textbf{Hyperparameters} & \textbf{Value} \\
\midrule
\multirow{4}{*}{Max planning iteration} & 10 (low/medium), 20 (high) \\
 & {\small\textcolor{gray}{(Obstruction/Kinematic)}} \\[0.5em]
 & 15 (low/medium), 30 (high) \\
 & {\small\textcolor{gray}{(Combination)}} \\
\bottomrule
\end{tabular}
\caption{Statler hyperparameters}
\end{table*}

\paragraph{LLM-DM.} This method employs a neurosymbolic planning approach by generating PDDL (Planning Domain Definition Language) files via VLM and leverage symbolic solver to derive action plan. 
We adapt this framework for cross-domain demo-to-code by using VLMs to predict domain file from source demonstration and problem file from deployment information, then iteratively refine the domain file until a valid plan is found through PDDL solving and using the plan as specification to generate code policy.

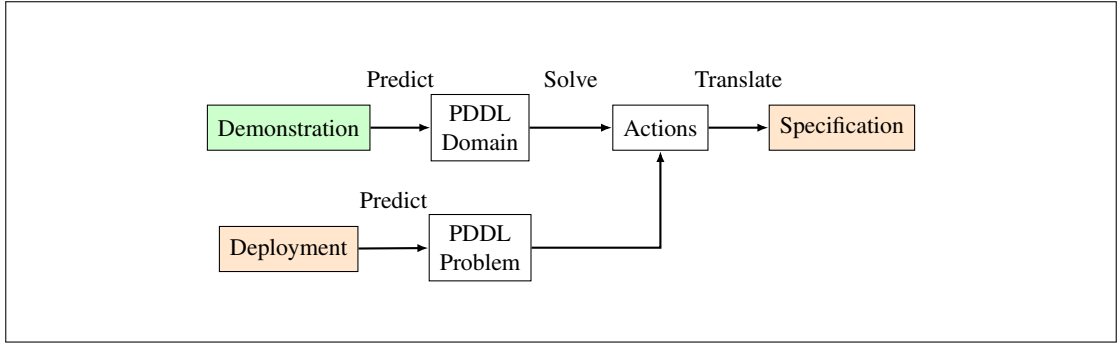
\begin{figure*}[h]
\centering
\begin{varwidth}{0.9\linewidth}
\adjustbox{padding=10mm 8mm 10mm 8mm, frame}{
\begin{minipage}{0.8\linewidth}
\centering
\begin{tikzpicture}[
scale=0.75,
node distance=6mm and 8mm,
every node/.style={font=\small},
box/.style={draw, minimum height=6mm, inner sep=4pt, align=center},
arrow/.style={-{Latex[length=1.5mm]}, thick}
]
\node[box, fill=green!20] (demo) {Demonstration};
\node[box, right=of demo] (domain) {PDDL\\Domain};
\node[box, fill=orange!20, below=10mm of demo] (scene) {Deployment};
\node[box, below=7mm of domain] (problem) {PDDL\\Problem};
\node[box, right=11mm of domain] (plan) {Actions};
\node[box, fill=orange!20, right=of plan] (spec) {Specification};

\draw[arrow] (demo) -- node[above=4mm] {Predict} (domain);
\draw[arrow] (scene) -- node[above=4mm] {Predict} (problem);
\draw[arrow] (domain) -- node[above=4mm] {Solve} (plan);
\draw[arrow] (problem) -| (plan);
\draw[arrow] (plan) -- node[above=4mm] {Translate} (spec);
\end{tikzpicture}
\end{minipage}
}
\end{varwidth}
\caption{LLM-DM high-level flow}
\end{figure*}

We implement LLM-DM referring to the official repository~\footnote{https://github.com/GuanSuns/LLMs-World-Models-for-Planning}, with our own implementation for the problem file prediction.
Implementation of the adapted LLM-DM consists of: (1) \textit{domain prediction}, where VLMs analyze source demonstrations to propose additional predicates beyond those in \ours, along with actions, for constructing a PDDL domain file; (2) \textit{problem prediction}, where VLMs observe the deployment scene to formulate a PDDL problem file using the proposed predicates; (3) \textit{PDDL-based solving}, where a symbolic solver attempts to synthesize an action sequence satisfying the given files; and (4) \textit{domain refinement}, where upon solving failure, VLMs analyze the failure and refine the domain model. 
This plan-refine cycle repeats until a valid plan is found or maximum attempts are reached.
For the symbolic solver, we use Fast Downward as provided in the official PDDLGym Planners repository~\footnote{https://github.com/ronuchit/pddlgym\_planners}.

\begin{table*}[htbp]
\centering
\label{app:hyper:llmdm}
\begin{tabular}{c c}
\toprule
\textbf{Hyperparameters} & \textbf{Value} \\
\midrule
\multirow{4}{*}{Max domain refinement} & 10 (low/medium), 20 (high) \\
 & {\small\textcolor{gray}{(Obstruction/Kinematic)}} \\[0.5em]
 & 15 (low/medium), 30 (high) \\
 & {\small\textcolor{gray}{(Combination)}} \\
\bottomrule
\end{tabular}
\caption{LLM-DM hyperparameters}
\end{table*}

\clearpage

\section{Additional Experiments}

\subsection{Experiments on VLM State Translation}

To evaluate the performance of symbolic state translation, we conduct experiments following \cite{sggeval}, with results presented in Table~\ref{tab:supp:vlm_sg}. The results show that the VLM reliably translates raw frames into scene graphs, achieving an F1-score above 0.82 against the ground-truth symbolic states of real-world scenes using GPT-5.

\begin{table*}[h]
\begin{center}
\begin{small}
\begin{adjustbox}{width=0.35\linewidth}
\begin{tabular}{l | ccc}
    \toprule
    \multirow{2}{*}{Method}
    & \multicolumn{3}{c}{\textit{Real-world}} \\
    \cmidrule(rl){2-4}
    & Precision & Recall & F1  \\
    \midrule
    \multicolumn{4}{l}{\textbf{Method}: $\ours$} \\
    \midrule
    GPT-5 
    & 0.97
    & 0.71
    & 0.82\\
    GPT-5-mini 
    & 0.88
    & 0.75
    & 0.80\\
    \bottomrule
\end{tabular}
\end{adjustbox}
\end{small}
\end{center}
\vspace{-10pt}
\caption{
Performance on VLM symbolic state translation
}
\label{tab:supp:vlm_sg}
\end{table*}

\subsection{Ablations on VLM choice}
Table~\ref{tab:main:vlm_choice} reports the ablation study on VLM choice for $\ours$. 
The results show that while GPT-5 achieves the best performance (with 80.00\% SR, 88.33\% GC), smaller models like GPT-5-mini and GPT-4 series still maintain competitive results with SR ranging from 63-70\% and GC above 75\%. 
This shows that $\ours$ remains effective across varying model scales.

\begin{table*}[h]
\begin{center}
\begin{small}
\begin{adjustbox}{width=0.5\linewidth}
\begin{tabular}{l | ccc}
    \toprule
    \multirow{2}{*}{Method}
    & \multicolumn{3}{c}{\textit{Combined}}  \\
    \cmidrule(rl){2-4}
    & SR  & GC  & PD   \\
    \midrule
    \multicolumn{4}{l}{\textbf{Method}: $\ours$} \\
    \midrule
    GPT-5 
    & 80.00 $\pm$7.43 
    & 88.33 $\pm$4.60 
    & 3.75 $\pm$2.61 \\
    \rowcolor{lightgray!30}
    GPT-5-mini
    & 70.00 $\pm$8.51 
    & 76.67 $\pm$7.08 
    & 11.43 $\pm$4.04 \\
    GPT-4.1
    & 66.67 $\pm$8.75 
    & 80.00 $\pm$5.67 
    & 9.00 $\pm$4.16 \\
    \rowcolor{lightgray!30}
    GPT-4o
    & 63.33 $\pm$8.95 
    & 75.00 $\pm$6.67 
    & 11.05 $\pm$4.39 \\
    \bottomrule
\end{tabular}
\end{adjustbox}
\end{small}
\end{center}
\vspace{-10pt}
\caption{
Ablation on VLM choice
}
\label{tab:main:vlm_choice}
\end{table*}

\subsection{Robustness to Scene Graph Perturbation}
\label{sec:supp:robustness}

To evaluate the robustness of \ours~to imperfect symbolic state translations, we conduct experiments under two types of scene graph perturbations: (1) randomly dropping 10\% of relations, and (2) injecting 10\% noisy (incorrect) relations.
As shown in Table~\ref{tab:supp:robustness}, \ours~maintains stable performance even when perturbations are applied to all scene graphs along the demonstration.
This robustness stems from the fact that perturbation-induced precondition violations are handled identically to violations arising from true cross-domain mismatches, through the same verification and refinement loops employed in both the symbolic world modeling and adaptation phases.

\begin{table}[h]
\begin{center}
\begin{small}
\begin{adjustbox}{width=0.45\linewidth}
\begin{tabular}{l | ccc}
    \toprule
    & None & 10\% Drop & 10\% Noise \\
    \midrule
    \ours & 65.00$\pm$7.64 & 55.00$\pm$7.97 & 60.00$\pm$7.84 \\
    \bottomrule
\end{tabular}
\end{adjustbox}
\end{small}
\end{center}
\vspace{-10pt}
\caption{
Robustness to scene graph perturbation
}
\label{tab:supp:robustness}
\end{table}

\subsection{Running Time Analysis}
\label{sec:supp:runtime}

Table~\ref{tab:supp:runtime} reports the running time comparison between iterative methods, decomposed into VLM inference time and symbolic tool execution time.
\ours~incurs minimal overhead compared to other iterative baselines, as symbolic tool operates as a forward executor that performs sequential state transitions with complexity linear in the demonstration length, rather than exponential symbolic search.
The primary computational bottleneck is VLM inference, a cost shared across all iterative baselines; the symbolic tool itself contributes marginal latency.

\begin{table}[h]
\begin{center}
\begin{small}
\begin{adjustbox}{width=0.45\linewidth}
\begin{tabular}{l | ccc}
    \toprule
    Method & VLM & Symbolic & Total \\
    \midrule
    Critic-V & 196.23s & -- & 196.23s \\
    Statler & 83.96s & -- & 83.96s \\
    LLM-DM & 91.56s & 0.40s & 91.96s \\
    \ours & 118.22s & 0.01s & 118.23s \\
    \bottomrule
\end{tabular}
\end{adjustbox}
\end{small}
\end{center}
\vspace{-10pt}
\caption{
Running time comparison between iterative methods.
}
\label{tab:supp:runtime}
\end{table}

\clearpage
\section{Additional Visualizations}

\subsection{Figure 5 in Main Paper Visualizations}
\begin{figure*}[h]
\begin{center}
\includegraphics[width=0.95\linewidth]{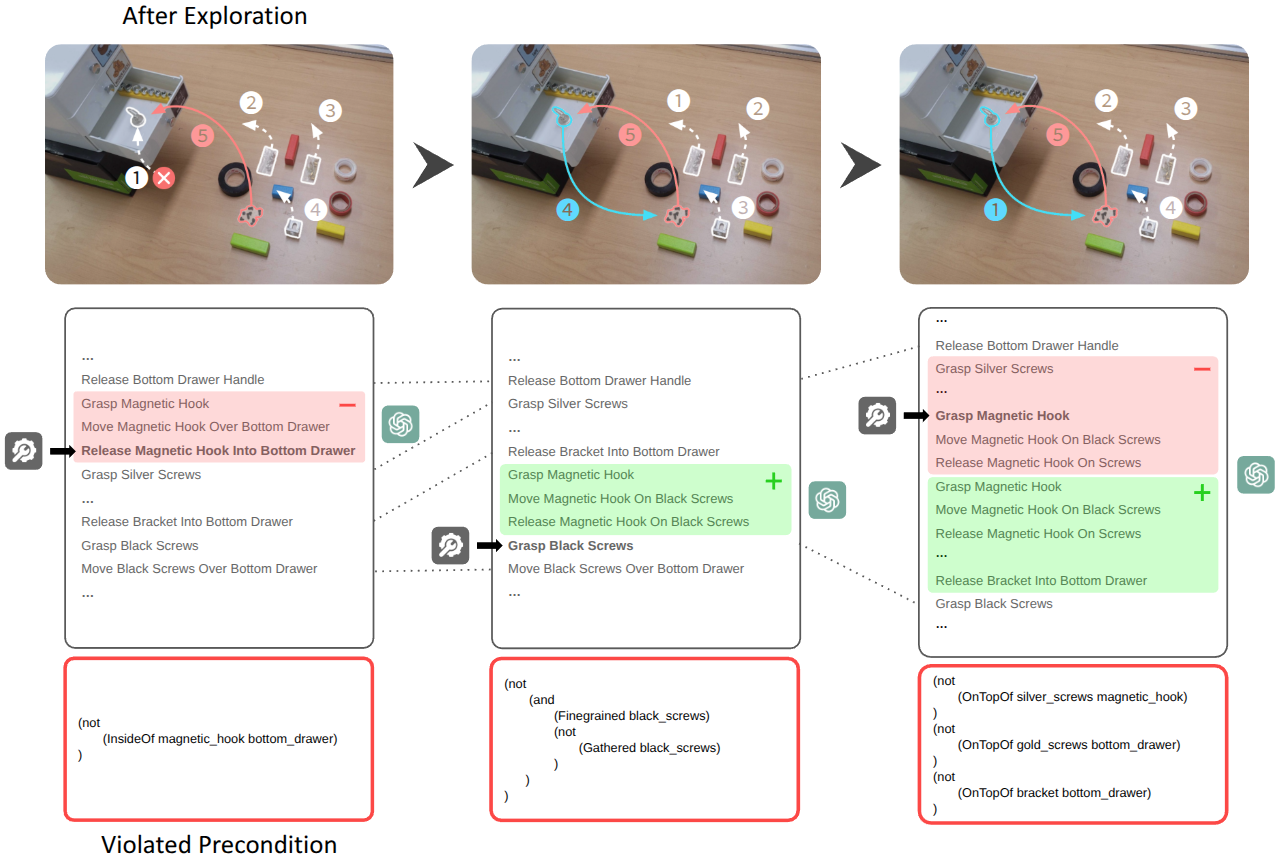}
\end{center}
\vspace{-10pt}
\caption{An expanded visualization of the adaptation process illustrated in Figure 1}
\label{fig:vis_diff}
\end{figure*}

\begin{figure*}[h]
\begin{center}
\includegraphics[width=0.80\linewidth]{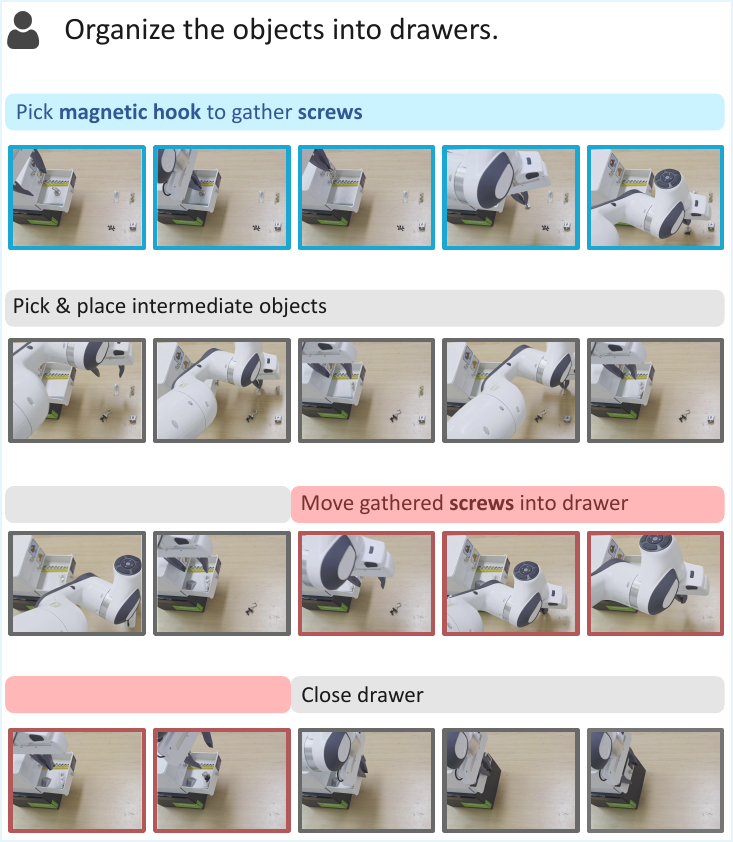}
\end{center}
\vspace{-10pt}
\caption{An expanded visualization of the task execution illustrated in Figure 1}
\label{fig:vis_main}
\end{figure*}

In this section, we illustrate the adaptation process of $\ours$ in Figure~\ref{fig:vis_diff}, and the execution of the generated code policy for our main scenario in Figure~\ref{fig:vis_main}.

\paragraph{Left.} The demonstration-derived procedure includes the redundant action ``Release Magnetic Hook Into Bottom Drawer'' since the action requires the target object to not be in the target position whereas magnetic hook already occupies its target position in the deployment domain, violating the precondition \texttt{(not (InsideOf magnetic\_hook bottom\_drawer))} identified by the symbolic tool.
Based on the violated precondition and the current state showing that the magnetic hook is already in place, the VLM generates an exploration to remove the redundant action chunk.
\\
\paragraph{Middle.} A domain gap exists in the deployment domain where the fine-grained object, black screws are scattered rather than gathered, violating the precondition for action ``Grasp Black Screws'' \texttt{(not (and (Finegrained black\_screws) (not (Gathered black\_screws))))}.
Based on the violated precondition and current state showing that magnetic hook is present in the scene, the VLM generates an exploration that repurposes the magnetic hook as an aggregation tool, inserting new actions to collect the scattered screws before attempting to grasp them.
\\
\paragraph{Right.} ``Grasp Magnetic Hook'' in the newly added hook retrieval action causes a precondition violation \texttt{(not (and (OnTopOf silver\_screws magnetic\_hook) (not (OnTopOf gold\_screws bottom\_drawer)) (not (OnTopOf bracket bottom\_drawer))))} indicating that multiple objects are stacked on top of the hook. 
VLM resolves by reordering the procedure: the hook retrieval and screw aggregation steps are moved earlier in the sequence, before organizing intermediate objects.
This resolves the violation, producing a final procedure compatible with the deployment domain.

\clearpage

\clearpage

\subsection{Real-world Experiment Visualizations}

\begin{figure*}[h]
\begin{center}
\includegraphics[width=0.69\linewidth]{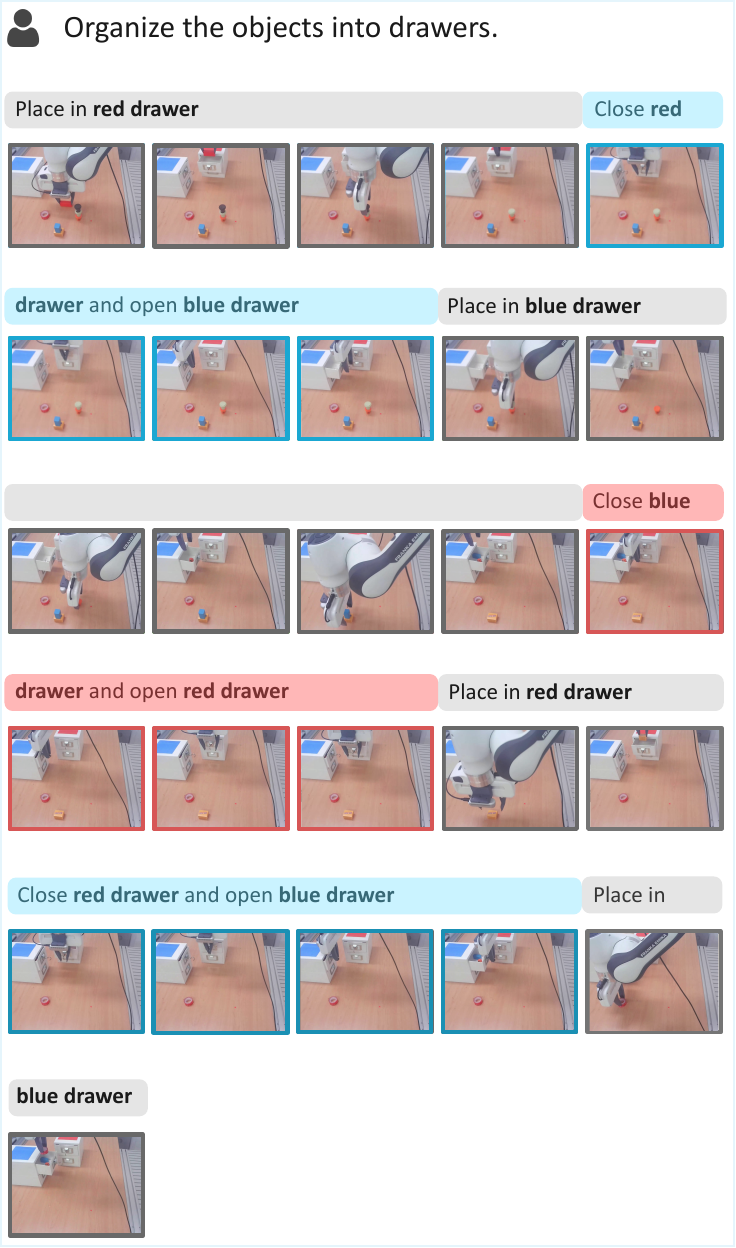}
\end{center}
\vspace{-10pt}
\caption{Visualization of real-world experiment in Table 2}
\label{fig:vis_realworld}
\vspace{-10pt}
\end{figure*}

In this section, we illustrate the execution of the generated code policy for our real-world experiment in Figure~\ref{fig:vis_realworld}.

\section{Prompts}
\subsection{Demo2Code}
\begin{tcolorbox}[
    colback=lightgray,
    colframe=darkgray,
    coltitle=white,
    title=Demo2Code | Observation Prediction,
    fonttitle=\bfseries,
    arc=1mm,
    breakable
]
\textbf{\large \texttt{[System]}}  

You are a scene descriptor for robotics. 
The user will provide three synchronized images of the scene (top/front/back views).
Your task is to describe what is happening in the scene and how are objects positioned and oriented based on provided scene and additional information.

\medskip

\textbf{Output requirements:}

\begin{itemize}
    {\leftskip=1em
    \item Provide one clear and concise scene description including the position and orientation of the objects.
    \item Start the scene description with an unordered dash (-).
    \par}
\end{itemize}

\tcblower
\textbf{\large \texttt{[User]}}

 Your task is to describe what is happening in the scene and how are objects positioned and oriented based on provided scene and additional information.

\medskip
Below is an example. \\ \texttt{[Example]}

\medskip

Now provide the scene descriptions for the following scene and information.

\medskip

\textbf{Information}
\begin{itemize}
    {\leftskip=1em
    \item Instruction: \\ \texttt{\{instruction\}} \medskip
    \item Object in scene: \\ \texttt{\{objects\}} \medskip
    \item Object state in scene: \\ \texttt{\{object\_state\}} \medskip
    \par}
\end{itemize}
\medskip

\textbf{Observation Description}
\end{tcolorbox}


\begin{tcolorbox}[
    colback=lightgray,
    colframe=darkgray,
    coltitle=white,
    title=Demo2Code | Action Prediction,
    fonttitle=\bfseries,
    arc=1mm,
    breakable
]
\textbf{\large \texttt{[System]}}

You are an action predictor for robotics. 
The user will provide observation sequence consisting of scene descriptions for each timestep and additional scene information.
Your task is to infer the high-level action that occurs between each pair of consecutive timesteps based on provided observation sequence and scene information.

\medskip

\textbf{Output requirements:}

\begin{itemize}
    {\leftskip=1em
    \item Provide one clear and concise action description, including the semantic and movement details of objects, for each transition between timesteps.
    \item Start the action description with ordered numbering (1., 2., ...).
    \item Use scene information and frames to identify the possible inconsistencies between observations and the scene, if exist infer the adapted actions.
    \par}
\end{itemize}

\tcblower
\textbf{\large \texttt{[User]}}

Your task is to infer the high-level action that occurs between each pair of consecutive timesteps based on provided observation sequence and scene information.

\medskip

Below are the definitions of the predicates. \\
\{predicates\}

\medskip

Below is an example. \\ \texttt{[Example]}

\medskip

Now provide the action descriptions for the following observation sequence and scene information.

\medskip

\textbf{Information}

\begin{itemize}
    {\leftskip=1em
    \item Instruction: \\ \texttt{\{instruction\}} \medskip 
    \item Object in scene: \\ \texttt{\{objects\}} \medskip 
    \item Object state in scene: \\ \texttt{\{object\_state\}} \medskip 
    \item Observation Descriptions: \\ \texttt{\{observations\}} \medskip 
    \par}
\end{itemize}
\medskip

\textbf{Action Descriptions}

\end{tcolorbox}


\subsection{GPT4V-Robotics}
\begin{tcolorbox}[
    colback=lightgray,
    colframe=darkgray,
    coltitle=white,
    title=GPT4V-Robotics | Domain Description,
    fonttitle=\bfseries,
    fontupper=\normalsize,
    arc=1mm,
    breakable
]
\textbf{\large \texttt{[System]}} \\  
You are a domain descriptor for robotics.
The user will provide three synchronized image sequences of the demonstration (top/front/back views).
Your task is to describe the domain and goal of the task being demonstrated in single-line natural language based on provided demonstration and information. \\ \\
\textbf{Output requirements:} 
\begin{itemize}
    {\leftskip=1em
    \item Produce one clear and specific domain description which can help future task planning for same domain in new scenes, including the specific goal information. 
    \item Start and end the domain description with triple backticks (\textasciigrave\textasciigrave\textasciigrave).
    \par}
\end{itemize}

\tcblower
\textbf{\large \texttt{[User]}}

Your task is to describe the domain and goal of the task being demonstrated in single-line natural language based on provided demonstration and information. \\ 

Below is an example. \\ 
\texttt{[Example]} \\

Now provide the domain descriptions for the following demonstration and information. \\ 

\textbf{Information}
\begin{itemize}
    {\leftskip=1em
    \item Instruction: \\ \texttt{\{instruction\}} \medskip 
    \item Object in scene:  \\ \texttt{\{objects\}} \medskip 
    \item Object state for each timestep:  \\ \texttt{\{object\_state\}}
    \par}
\end{itemize} 

\medskip
\medskip
\textbf{Domain Description}

\end{tcolorbox}


\begin{tcolorbox}[
    colback=lightgray,
    colframe=darkgray,
    coltitle=white,
    title=GPT4V-Robotics | Action Planning,
    fonttitle=\bfseries,
    arc=1mm,
    breakable
]
\textbf{\large \texttt{[System]}} \\ 
You are an task planner for robotics.
The user will provide three synchronized images of the scene (top/front/back views).
Your goal is to generate task plan for scene based on provided task information and deployment scene.

\medskip

\textbf{Output requirements:}

\begin{itemize}
    {\leftskip=1em
    \item Provide a step-by-step action plan to accomplish the task with each clear and concise single-line actions.
    \item Start the action plan with ordered numbering (1., 2., ...).
    \par}
\end{itemize}

\tcblower
\textbf{\large \texttt{[User]}}

Your goal is to generate task plan for scene based on provided task information and deployment scene.

\medskip
Below are the definitions of the predicates. \\
\{predicates\}

\medskip
Below is an example. \\ \texttt{[Example]}

\medskip

 Now provide the action plan for the following information.

\medskip

\textbf{Information}
\begin{itemize}
    {\leftskip=1em
    \item Domain Description: \\ \texttt{\{domain\_description\}}  \medskip 
    \item Instruction: \\ \texttt{\{instruction\}} \medskip 
    \item Objects in scene: \\ \texttt{\{objects\}}  \medskip 
    \item Gripper state in scene: \\ \texttt{\{gripper\_state\}} \medskip 
    \par}
\end{itemize}
\medskip
\medskip
\textbf{Action Plan}
\end{tcolorbox}

\subsection{Critic-V}
\begin{tcolorbox}[
    colback=lightgray,
    colframe=darkgray,
    coltitle=white,
    title=Critic-V | Feedback Generation,
    fonttitle=\bfseries,
    arc=1mm,
    breakable
]
\textbf{\large \texttt{[System]}} 

You are a feedback generator for robotics.
The user will provide source action plan with task information and three synchronized images of the deployment scene (top/front/back views).
Your task is to analyze the source action plan and give a feedback on how to refine it to succeed on the task in the deployment scene.

\medskip

\textbf{Output requirements:}

\begin{itemize}
    {\leftskip=1em
    \item Provide clear and concise feedback on how to correct the action plan for deployment scene, if no problem exists, state \texttt{``No issues''}.
    \item Only provide feedback for the most major critical issue that would lead to task failure in the deployment scene.
    \item Start the feedback with an unordered dash (-).
    \par}
\end{itemize}

\tcblower
\textbf{\large \texttt{[User]}}

Your task is to analyze the source action plan and give a feedback on how to refine it to succeed on the task in the deployment scene.

\medskip

Below is an example. \\ \texttt{[Example]}

\medskip

Now provide the feedback for the following initial action plan and information.

\medskip

\textbf{Information}

\begin{itemize}
{\leftskip=1em
    \item Instruction: \\ \texttt{\{instruction\}} \medskip
    \item Objects in scene: \\ \texttt{\{objects\}} \medskip
    \item Object state in scene: \\ \texttt{\{object\_state\}} \medskip
    \item Initial Action Plan: \\ \texttt{\{demo\_summary\}} \medskip
\par}
\end{itemize}

\textbf{Correction Feedback}

\medskip

\end{tcolorbox}


\begin{tcolorbox}[
    colback=lightgray,
    colframe=darkgray,
    coltitle=white,
    title=Critic-V | Correction Proposal,
    fonttitle=\bfseries,
    arc=1mm,
    breakable
]
\textbf{\large \texttt{[System]}}

You are an action plan corrector for robotics.
The user will provide an action plan with a feedback for better task success.
Your task is to generate a \texttt{SEARCH/REPLACE} patch that applies the feedback to the action plan.

\medskip

\textbf{Output requirements:}

\begin{itemize}
    {\leftskip=1em
    \item Use commonsense reasoning to propose a patch that applies the feedback, ensuring the action plan is executable in the deployment scene.
    \item First provide a reasoning about the root cause of the feedback, how to realize the feedback as an actual action patch, and why your proposed patch works.
    \item Propose one \texttt{SEARCH} block and one \texttt{REPLACE} block which can be applied to the original action plan to apply the feedback.
    \item Format your response in the following way:
    \par}
\end{itemize}

\medskip

\textbf{Information}

\begin{itemize}
{\leftskip=1em
\item (your reasoning here starting with a dash)
\par}
\end{itemize}

\medskip

\textbf{Correction patch:}

{\leftskip=2em
    \begin{verbatim}
    <<<<<<< SEARCH
    ActionToRemove_1
    ActionToRemove_2
    ...
    =======
    ActionToAdd_1
    ActionToAdd_2
    ...
    >>>>>>> REPLACE
    \end{verbatim}
\par}

\tcblower

\textbf{\large \texttt{[User]}}

Your task is to generate a \texttt{SEARCH/REPLACE} patch that applies the feedback to the action plan.

\medskip

Below is an example. \\ \texttt{[Example]}

\medskip

Now provide the correction patch for the following action plan and feedbacks.

\medskip

\textbf{Information}

\begin{itemize}
    {\leftskip=1em
    \item Instruction: \\ \texttt{\{instruction\}} \medskip
    \item Objects in scene: \\ \texttt{\{objects\}} \medskip
    \item Action Plan: \\ \texttt{\{demo\_summary\}} \medskip
    \item Feedback: \\ \texttt{\{feedback\}} \medskip
    \par}
\end{itemize}
\medskip

\textbf{Correction rationale}

\medskip

\end{tcolorbox}



\subsection{MoReVQA}
\begin{tcolorbox}[
    colback=lightgray,
    colframe=darkgray,
    coltitle=white,
    title=MoReVQA | Event Parsing,
    fonttitle=\bfseries,
    arc=1mm,
    breakable
]

\textbf{\large \texttt{[System]}}

You are a parsed event maker to answer the plans of the question for event parsing.
The user will provide a question.

\medskip

\textbf{Output requirements:}

\begin{itemize}
    {\leftskip=1em
    \item[] 1. question: Change the instruction into a question. \\
    2. conjunction: One of (And / Or / None). \\ 
    3. parse\_event: A list of sub-events split by conjunction. If none exists, include one event. \\
    4. event\_object: Specify the main object(s) for each sub-event. \\ 
    5. classify: One of (which / where / why / how).
    \par}
\end{itemize}

\tcblower

\textbf{\large \texttt{[User]}}

You are a parsed event maker to answer the plans of the question for event parsing.

\medskip

Below is an example. \\ \texttt{[Example]}

\medskip

Now make the Parsed Event based on the provided information.

\medskip

\textbf{Information}

\begin{itemize}
    {\leftskip=1em
    \item Instruction: \\ \texttt{\{instruction\}} \medskip
    \par}
\end{itemize}

\medskip

\textbf{Answer}

\end{tcolorbox}


\begin{tcolorbox}[
    colback=lightgray,
    colframe=darkgray,
    coltitle=white,
    title=MoReVQA | Event Grounding,
    fonttitle=\bfseries,
    arc=1mm,
    breakable
]

\textbf{\large \texttt{[System]}}

You are a grounding module that aligns parsed events with the visual frames of a target scene.
The user will provide feedback (optional), domain description, target scene, target objects, parsed events, and event objects.
Your task is to create specific parsed events and objects grounded in the given scene and target objects, ensuring physical feasibility.

\medskip

\textbf{Output requirements:} JSON format.

\begin{itemize}
    {\leftskip=1em
    \item \texttt{parse\_event}: scene-grounded, physically feasible.
    \item \texttt{event\_object}: specify subject and target for each grounded event.
    \par}
\end{itemize}

\tcblower

\textbf{\large \texttt{[User]}}

You are a grounding module to align parsed events with the visual frames of a target scene.

\medskip

Below is an example. \\ \texttt{[Example]}

\medskip

Now make specific parsed events and event objects.

\medskip

\textbf{Information}

\begin{itemize}
    {\leftskip=1em
    \item Domain Description: \\ \texttt {\{domain\_description\}} \medskip
    \item Target Objects: \\ \texttt {\{target\_objects\}} \medskip
    \item Event Queue: \\ \texttt {\{event\_queue\}} \medskip
    \item Event Object: \\ \texttt {\{event\_object\}}
    \par}
\end{itemize}
\medskip

\textbf{Specific parsed events and event objects}

\end{tcolorbox}


\begin{tcolorbox}[
    colback=lightgray,
    colframe=darkgray,
    coltitle=white,
    title=MoReVQA | M2 Verified API Generator,
    fonttitle=\bfseries,
    arc=1mm,
    breakable
]

\textbf{\large \texttt{[System]}}

You are an API generation module that verifies whether a given question is successfully executed.
Your task is to generate a verify API that determines whether the question can succeed.

\tcblower

\textbf{\large \texttt{[User]}}

You are an API generation module that verifies whether a given question is successfully executed.

\medskip

Below is an example. \\ \texttt{[Example]}

\medskip

Now generate the M2 verify API.

\medskip

\textbf{Information}

\begin{itemize}
    {\leftskip=1em
    \item Question: \\ \texttt{\{question\}} \medskip
    \par}
\end{itemize}

\medskip

\textbf{M2 verify API}

\end{tcolorbox}

\begin{tcolorbox}[
    colback=lightgray,
    colframe=darkgray,
    coltitle=white,
    title=MoReVQA | Verify API Executor,
    fonttitle=\bfseries,
    arc=1mm,
    breakable
]

\textbf{\large \texttt{[System]}}

You are a verify-API executor that determines whether the question can succeed if events are successfully executed.  
The user will provide target scene, event queue, and verify API.  
Your task is to verify whether the question itself can succeed assuming all events execute successfully.

\medskip

\textbf{Output requirements:} \\ JSON format.

\begin{itemize}
    {\leftskip=1em
    \item \texttt{event\_queue}: executed event queue.
    \item \texttt{verified}: true if question can succeed; plural or \texttt{``all''} still counts as true with one representative.
    \item \texttt{reason}: explanation if false; otherwise \texttt{``None''}.
    \item \texttt{verify\_action}: returns true if the question can succeed in target scene. \\
    \par}
\end{itemize}

\medskip

\tcblower

\textbf{\large \texttt{[User]}}

You are a verify-API executor that evaluates whether the given question can succeed.

\medskip

Below is an example. \\ \texttt{[Example]}

\medskip

Now generate the verified API.

\medskip

\textbf{Information}

\begin{itemize}
    {\leftskip=1em
    \item verify\_api: \\ \texttt{\{verify\_api\}} \medskip
    \item question: \\ \texttt{\{question\}} \medskip
    \item event\_queue: \\ \texttt{\{event\_queue\}} \medskip
    \par}
\end{itemize}
\medskip

\textbf{Result}

\end{tcolorbox}


\begin{tcolorbox}[
    colback=lightgray,
    colframe=darkgray,
    coltitle=white,
    title=MoReVQA | M3 VQA API Generation Module,
    fonttitle=\bfseries,
    arc=1mm,
    breakable
]

\textbf{\large \texttt{[System]}}

You are an M3 VQA API generation module.
Your task is to generate exactly three procedural VQA sub-questions for each event.

\medskip

\textbf{Assumptions:}

\begin{itemize}
    {\leftskip=1em
    \item All required objects exist in the scene, and the action is feasible.
    \item Do NOT ask existence questions.
    \item Focus on procedural HOW questions.
    \par}
\end{itemize}

\medskip

\textbf{Input:}

\begin{itemize}
    {\leftskip=1em
    \item question: event\_queue
    \par}
\end{itemize}

\medskip

\textbf{Output:}

\begin{itemize}[leftmargin=2em]
    \item For each event\_queue, output:
    \begin{itemize}[leftmargin=1em]
        \item One top-level VQA prompt phrased as \texttt{``How to <event>?''}
        \item Exactly three sub-questions specifying steps or constraints
    \end{itemize}
    \item Prefer imperative, scene-grounded, concise wording.
\end{itemize}
\medskip

\textbf{Output format:}

{\leftskip=1em
\verb|vqa("How to <event>?")|\newline
\verb|vqa(["<sub-q1>", "<sub-q2>", "<sub-q3>"])|
\par}

\tcblower

\textbf{\large \texttt{[User]}}

You are an M3 VQA API generation module that generates exactly three procedural VQA sub-questions.

\medskip

Below is an example. \\ \texttt{[Example]}

\medskip

Now generate the M3 VQA API.

\medskip

\textbf{Information}

\begin{itemize}     
    {\leftskip=1em
    \item Event Queue: \\ \texttt{\{event\_queue\}} \medskip
    \par}
\end{itemize}
\medskip

\textbf{M3 VQA API}

\end{tcolorbox}


\begin{tcolorbox}[
    colback=lightgray,
    colframe=darkgray,
    coltitle=white,
    title=MoReVQA | VQA API Executor,
    fonttitle=\bfseries,
    arc=1mm,
    breakable
]

\textbf{\large \texttt{[System]}} 

You are a VQA API executor that answers questions about the target scene
to evaluate whether demonstrated actions can be successfully reproduced.

\medskip

The user will provide:
\begin{itemize}
    {\leftskip=1em
    \item target\_scene
    \item event queue and event objects
    \par}
\end{itemize}

\medskip

\textbf{Output:} Answer the VQA.

\tcblower
\textbf{\large \texttt{[User]}}

You are a VQA API executor that evaluates whether the demonstrated actions
can be successfully reproduced in the given target scene.

\medskip

Below is an example: \\ \texttt[Example]

\medskip

Now make the answer for the VQA.

\medskip

\textbf{Information}

\begin{itemize}
    {\leftskip=1em
    \item VQA: \\ \texttt{\{vqa\}}
    \par}
\end{itemize}
\medskip

\textbf{Result}

\end{tcolorbox}


\begin{tcolorbox}[
    colback=lightgray,
    colframe=darkgray,
    coltitle=white,
    title=MoReVQA | Action Planning,
    fonttitle=\bfseries,
    arc=1mm,
    breakable
]

\textbf{\large \texttt{[System]}}

You are a task planner for robotics.
The user will provide three synchronized images of the scene.
Your goal is to generate task plans based on task information and the deployment scene.

\medskip

\textbf{Output requirements:}

\begin{itemize}
    {\leftskip=1em
    \item Provide a step-by-step action plan.
    \item Each action must be a clear and concise single-line instruction.
    \item Start with ordered numbering (1., 2., ...).
    \par}
\end{itemize}

\tcblower
\textbf{\large \texttt{[User]}}

Your goal is to generate an action plan for the scene based on the provided task info and deployment scene.

\medskip

Below are the definitions of the predicates: \\ \texttt{\{predicates\}}

\medskip

Below is an example: \\ \texttt{[Example]}

\medskip

Now provide the action plan for the following information.

\medskip

\textbf{Information}

\begin{itemize}
    {\leftskip=1em
    \item Event Queue: \\ \texttt{\{event\_queue\}} \medskip
    \item Objects in scene: \\ \texttt{\{target\_objects\}} \medskip
    \item Object state in scene: \\ \texttt{\{target\_object\_state\}} \medskip
    \item VQA Answer: \\ \texttt{\{vqa\_answer\}} \medskip
    \par}
\end{itemize}

\medskip

\textbf{Action Plan}

\end{tcolorbox}


\subsection{Statler}
\begin{tcolorbox}[
    colback=lightgray,
    colframe=darkgray,
    coltitle=white,
    title=Statler | Action Planning,
    fonttitle=\bfseries,
    arc=1mm,
    breakable
]

\textbf{\large \texttt{[System]}}

You are a task planner for robotics.
The user will provide three synchronized images of the scene (top/front/back views).
Your goal is to continue the task plan with a few actions and predict the state after the plan
based on provided task information and the deployment scene.

\medskip

\textbf{Output requirements:}

\begin{itemize}
    {\leftskip=1em
    \item Provide a step-by-step partial action plan (at most 10 steps) to head toward the goal; if the goal is reached, output \texttt{``Goal reached''}.
    \item Start the action plan with ordered numbering (1., 2., ...).
    \item After the action plan, output the predicted state as a list of grounded atoms holding true for the scene after executing the action plan.
    \item Start each grounded atom of the state with an unordered dash (-).
    \par}
\end{itemize}

\tcblower
\textbf{\large \texttt{[User]}}

Your goal is to continue the task plan with a few actions for the scene based on the provided task information and deployment scene.

\medskip

Below are the definitions of the predicates. \\ \texttt{\{predicates\}}

\medskip

Below is an example. \\ \texttt{[Example]}

\medskip

Now use the same format for the following information.

\medskip

\textbf{Information}

\begin{itemize}
    {\leftskip=1em
    \item Domain Description: \\ \texttt{\{domain\_decription\}} \medskip
    \item Instruction: \\ \texttt{\{instruction\}} \medskip
    \item Object in scene: \\ \texttt{\{objects\}} \medskip
    \par}
\end{itemize}
\medskip

\textbf{Previous Actions and Current State}

\begin{itemize}
    {\leftskip=1em
    \item Previous Actions: \\ \texttt{\{demo\_summary\}} \medskip
    \item Current State: \\ \texttt{\{current\_state\}} \medskip
    \par}
\end{itemize}

\medskip

\textbf{Action Plan (at most 10 steps)}

\medskip

\textbf{Predicted State}

\end{tcolorbox}


\subsection{LLM-DM}
\begin{tcolorbox}[
    colback=lightgray,
    colframe=darkgray,
    coltitle=white,
    title=LLM-DM | Action Recommendation,
    fonttitle=\bfseries,
    arc=1mm,
    breakable
]

\textbf{\large \texttt{[System]}}

You are a PDDL action recommender for robotics. 
The user will provide a description of the domain and task instruction, along with the objects in the scene.
Your task is to recommend the useful PDDL actions to solve the task in natural language.

\medskip

\textbf{Output requirements:}

\begin{itemize}
    {\leftskip=1em
    \item The actions should have a name and a short description of what the action does in natural language.
    \item Start the actions with ordered numbering (1., 2., ...).
    \item Make the actions general enough to be reusable in different tasks within the same domain.
    \par}
\end{itemize}

\tcblower
\textbf{\large \texttt{[User]}}

Your task is to recommend the useful PDDL actions to solve the task in natural language.

\medskip

Below is an example. \\ \texttt{[Example]}

\medskip

Now recommend the actions based on the provided demonstration information.

\medskip

\textbf{Demonstration Information}

\begin{itemize}
    {\leftskip=1em
    \item Domain Description: \\ \texttt{\{domain\_description\}} \medskip
    \item Instruction: \\ \texttt{\{instruction\}} \medskip
    \item Objects in scene: \\ \texttt{\{objects\}} \medskip
    \par}
\end{itemize}
\medskip

\textbf{Action Recommendations}

\end{tcolorbox}


\begin{tcolorbox}[
    colback=lightgray,
    colframe=darkgray,
    coltitle=white,
    title=LLM-DM | Predicate Proposal,
    fonttitle=\bfseries,
    arc=1mm,
    breakable
]

\textbf{\large \texttt{[System]}}

You are a predicate proposer for robotics.
The user will provide target initial state, base predicates, and domain/action descriptions in natural language.
Your task is to propose a set of untyped predicates with associated descriptions that is useful to define the actions in PDDL format.

\medskip

\textbf{Output requirements:}

\begin{itemize}
    {\leftskip=1em
    \item Provide one clear and specific list of untyped predicates with descriptions that is useful for defining the actions in PDDL format.
    \item Start the predicates with unordered dash (-).
    \par}
\end{itemize}

\tcblower
\textbf{\large \texttt{[User]}}

Your task is to propose a set of untyped predicates with associated descriptions that is useful to define the actions in PDDL format.

\medskip

Below are the definitions of the existing predicates. Do not invent predicates with duplicated meanings. \\ \texttt{\{predicates\}}

\medskip

Below is an example. \\ \texttt{[Example]}

\medskip

Now propose predicates based on the provided domain description and action descriptions.

\medskip

\textbf{Domain Description}

\begin{itemize}
    {\leftskip=1em
    \item \texttt{\{domain\_description\}}
    \par}
\end{itemize}

\medskip

\textbf{Action Descriptions}

\begin{itemize}
    {\leftskip=1em
    \item \texttt{\{action\_description\}}
    \par}
\end{itemize}

\medskip

\textbf{Predicate Proposal}

\end{tcolorbox}


\begin{tcolorbox}[
    colback=lightgray,
    colframe=darkgray,
    coltitle=white,
    title=LLM-DM | Action Construction,
    fonttitle=\bfseries,
    arc=1mm,
    breakable
]

\textbf{\large \texttt{[System]}}

You are an action constructor for robotics. 
The user will provide action description in natural language and predicates. 
Your task is to convert the action description into a PDDL-style action definition based on the provided domain description and predicate set.

\medskip

\textbf{Output requirements:}

\begin{itemize}
    {\leftskip=1em
    \item Produce one clear PDDL-style action definition for the action in the action description.
    \item The action definition should include the action name, parameters, preconditions, and effects.
    \item Start the name with \texttt{``-''} (dash followed by a space) under the \textbf{Name:} section; the action name should not overlap with predicate names.
    \item Start the parameters with ordered numbering (1., 2., ...) under the \textbf{Parameters:} section.
    \item Start and end the preconditions and effects with triple backticks (\textasciigrave\textasciigrave\textasciigrave) under the \textbf{Preconditions:} and \textbf{Effects:} sections respectively.
    \item All predicates used in the action definition must be from the provided predicates.
    \item Use predicate names exactly as given in the provided predicate list; do not invent new predicates or rename/alias any predicate.
    \item You can use \texttt{(not ...)}, \texttt{(and ...)} in the preconditions and effects; do not use \texttt{(or ...)}, \texttt{(when ...)}.
    \par}
\end{itemize}

\tcblower
\textbf{\large \texttt{[User]}}

Your task is to convert the action description into a PDDL-style action definition based on the provided domain description and predicate set.

\medskip

Below is an example. \\ \texttt{[Example]}

\medskip

Now convert the action description into PDDL-style action definition.

\medskip

\textbf{Action Description}

\begin{itemize}
    {\leftskip=1em
    \item \texttt{\{action\_description\}}
    \par}
\end{itemize}

\medskip

\textbf{Predicates}

\begin{itemize}
    {\leftskip=1em
    \item \texttt{\{predicates\}}
    \par}
\end{itemize}

\medskip

\textbf{Action Definition}

\end{tcolorbox}

\begin{tcolorbox}[
    colback=lightgray,
    colframe=darkgray,
    coltitle=white,
    title=LLM-DM | Problem Prediction,
    fonttitle=\bfseries,
    arc=1mm,
    breakable
]

\textbf{\large \texttt{[System]}}

You are a PDDL problem predictor for robotics. 
The user will provide a domain description, a list of objects, a predicate set, and three synchronized images of the deployment scene (top/front/back views).
Your task is to predict the PDDL-style problem definition for the target initial scene.

\medskip

\textbf{Output requirements:}
\begin{itemize}
    {\leftskip=1em
    \item Produce one clear and specific PDDL-style problem definition for the target initial scene.
    \item The objects, initial state, and goal state should be in PDDL format separately wrapped in triple backticks (\textasciigrave\textasciigrave\textasciigrave) under \texttt{\textbf{``Objects:''}}, \texttt{\textbf{``Initial state:''}}, and \texttt{\textbf{``Goal state:''}} sections respectively.
    \item Start the content of objects, initial state, and goal state with \texttt{``(:objects''}, \texttt{``(:init''}, and \texttt{``(:goal''} respectively.
    \item All predicates in the initial state and goal state must be from the provided predicate set.
    \par}
\end{itemize}

\tcblower
\textbf{\large \texttt{[User]}}

Your task is to predict the PDDL-style problem definition for the target initial scene.

\medskip

Below is an example. \\ \texttt{[Example]}

\medskip

Now predict the PDDL-style problem definition based on the provided information.

\medskip

\textbf{Domain Information}

\begin{itemize}
    {\leftskip=1em
    \item Domain description: \\ \texttt{\{domain\_description\}} \medskip
    \item Predicates: \\ \texttt{\{predicates\}} \medskip
    \par}
\end{itemize}

\medskip

\textbf{Problem Information}

\begin{itemize}
    {\leftskip=1em
    \item Instruction: \\ \texttt{\{instruction\}} \medskip
    \item Objects in scene: \\ \texttt{\{objects\}} \medskip
    \item Object state in scene: \\ \texttt{\{object\_state\}} \medskip
    \par}
\end{itemize}

\medskip

\textbf{Predicted Problem Definition}

\end{tcolorbox}

\begin{tcolorbox}[
    colback=lightgray,
    colframe=darkgray,
    coltitle=white,
    title=LLM-DM | Domain Refinement,
    fonttitle=\bfseries,
    arc=1mm,
    breakable
]

\textbf{\large \texttt{[System]}}

You are a PDDL domain refiner for robotics.
The user will provide a problematic domain PDDL which failed to generate a plan for the given problem PDDL, and optionally some context about the failure if available.
Your task is to diagnose the domain PDDL based on the problem PDDL and apply all necessary fixes to make the problem solvable.

\medskip

\textbf{Output requirements:}

\begin{itemize}
    {\leftskip=1em
    \item First provide a reasoning about what is wrong with the original domain and how you fixed it.
    \item Start the refinement rationale with \texttt{``-''} (dash followed by a space) under the \texttt{\textbf{``Refinement Rationale:''}} section.
    \item Produce one refined domain PDDL that can solve the given problem PDDL; keep the name of the domain the same as the original.
    \item Start and end the refined domain PDDL with triple backticks (\textasciigrave\textasciigrave\textasciigrave) under the \texttt{\textbf{``Refined Domain PDDL:''}} section.
    \par}
\end{itemize}

\tcblower
\textbf{\large \texttt{[User]}}

Your task is to diagnose the domain PDDL based on the problem PDDL and apply all necessary fixes to make the problem solvable.

\medskip

Below are the definitions of the predicates. \\ \texttt{\{predicates\}}

\medskip

Now return the refined domain in a whole.

\medskip

\textbf{Domain PDDL}

\begin{itemize}
    {\leftskip=1em
    \item \texttt{\{domain\_pddl\}} \medskip
    \par}
\end{itemize}

\medskip

\textbf{Problem PDDL}

\begin{itemize}
    {\leftskip=1em
    \item \texttt{\{problem\_pddl\}} \medskip
    \par}
\end{itemize}

\medskip

\textbf{Context about failure (if any):}

\begin{itemize}
    {\leftskip=1em
    \item \texttt{\{failure\_context\}} \medskip
    \par}
\end{itemize}
\medskip

\textbf{Refinement Rationale:}

\medskip

\textbf{Refined Domain PDDL:}

\medskip

\end{tcolorbox}

\begin{tcolorbox}[
    colback=lightgray,
    colframe=darkgray,
    coltitle=white,
    title=LLM-DM | Plan Translation,
    fonttitle=\bfseries,
    arc=1mm,
    breakable
]

\textbf{\large \texttt{[System]}} 

You are a plan translator for robotics.
The user will provide a target plan in PDDL format.
Your task is to translate the plan into clear and concise natural language steps, preserving the order and intention of each action.

\medskip

\textbf{Output requirements:}

\begin{itemize}
    {\leftskip=1em
    \item Translate each action into natural language that describes what the robot is doing.
    \item Start the actions with ordered numbering (1., 2., ...).
    \par}
\end{itemize}

\tcblower
\textbf{\large \texttt{[User]}}

Your task is to translate the plan into clear, concise, human-readable task steps, preserving the order and intention of each action.

\medskip

Below is an example. \\ \texttt{[Example]}

\medskip

Now translate the PDDL-style plan into natural language based on the provided information.

\medskip

\textbf{Target plan:}

\begin{itemize}
    {\leftskip=1em
    \item \texttt{\{target\_plan\}} \medskip
    \par}
\end{itemize}

\medskip

\textbf{Natural-language Plan:}

\end{tcolorbox}


\subsection{NeSyCR}
\begin{tcolorbox}[
    colback=lightgray,
    colframe=darkgray,
    coltitle=white,
    title=NeSyCR | Action Prediction,
    fonttitle=\bfseries,
    arc=1mm,
    breakable
]

\textbf{\large \texttt{[System]}}

You are an action predictor for robotics.
The user will provide state transition information consisting of previous state, current state and state difference and scenes.
Your task is to predict the executed action's semantic, preconditions and effects based on provided state transition information.

\medskip

\textbf{Output requirements:}

\begin{itemize}
    {\leftskip=1em
    \item First provide a reasoning about what action was executed, why these preconditions were necessary, and why these effects occurred.
    \item Then produce one clear and specific action semantic inferred from the state transition.
    \item Start the action semantic with \texttt{``-''} (dash followed by a space).
    \item If preconditions is empty leave it as \texttt{``None''}. The effects can never be empty.
    \item Use negation in preconditions and effects when relevant (e.g., \texttt{(not (GripperHolding block))}).
    \item Use \texttt{forall} quantifiers when the preconditions or effects should apply to all possible objects (e.g., \texttt{(forall (?x - thing) (not (GripperHolding ?x)))}).
    \item Except the \texttt{forall} quantifier, all atoms must be fully grounded with specific object names.
    \par}
\end{itemize}

\tcblower
\textbf{\large \texttt{[User]}}

Your task is to predict the executed action's semantic, preconditions and effects based on provided state transition information.

\medskip

Below are the definitions of the predicates. \\ \texttt{\{predicates\}}

\medskip

Below are examples. \\ \texttt{[Example]}

\medskip

Now predict the action based on the following information.

\medskip

\textbf{State Description:}

\begin{itemize}
    {\leftskip=1em
    \item Instruction: \\ \texttt{\{instruction\}} \medskip
    \item Objects in scene: \\ \texttt{\{objects\}} \medskip
    \item Previous state (before action): \\ \texttt{\{prev\_state\}} \medskip
    \item Current state (after action): \\ \texttt{\{curr\_state\}} \medskip
    \item State difference: \\ \texttt{\{state\_diff\}}
    \par}
\end{itemize}

\medskip

\textbf{Prediction Rationale:}

\medskip

\textbf{Predicted Action:}

{\leftskip=1em
Action Semantic: \\[0.3em]
Precondition: \\[0.3em]
Effect:
\par}

\end{tcolorbox}

\begin{tcolorbox}[
    colback=lightgray,
    colframe=darkgray,
    coltitle=white,
    title=NeSyCR | Refinement Proposal,
    fonttitle=\bfseries,
    arc=1mm,
    breakable
]

\textbf{\large \texttt{[System]}}

You are an action plan refiner for robotics. 
Given the task information, action information and error details, propose a refinement patch so that the target state now meets the previously unmet preconditions of the erroneous action.

\medskip

\textbf{Output requirements:}

\begin{itemize}
    {\leftskip=1em
    \item Use commonsense reasoning to propose a patch that resolves the failure, ensuring the erroneous action's preconditions are satisfied.
    \item First provide a reasoning about the root cause of the error, how to make the erroneous action's preconditions satisfied, and why your proposed patch works.
    \item Propose one \texttt{SEARCH} block and one \texttt{REPLACE} block which can make preconditions of the erroneous action satisfied.
    \item The \texttt{SEARCH} block must contain a continuous, consecutive sequence of actions from the action plan in their exact form.
    \item Format your response in the following way:
    \par}
\end{itemize}

\medskip
    
{\leftskip=1em
    \textbf{Refinement rationale:} \\ - (your reasoning here starting with a dash) \\ 
    
    \medskip
    
    \textbf{Refinement patch:}

    \begin{verbatim}
    <<<<<<< SEARCH
    ActionToRemove_1
    - Preconditions: 
            ...
    - Effects: 
            ...
    
    ActionToRemove_2
    - Preconditions: 
            ...
    - Effects: 
            ...
    ...
    =======
    ActionToAdd_1
    - Preconditions: 
            ...
    - Effects: 
            ...
    
    ActionToAdd_2
    - Preconditions:
            ...
    - Effects:
            ...
    ...
    >>>>>>> REPLACE
    \end{verbatim}
    
\par}
\tcblower
\textbf{\large \texttt{[User]}}

Revise the action sequence with a \texttt{SEARCH/REPLACE} patch so unmet preconditions of the erroneous action are satisfied.

\medskip

Below are the predicate definitions: \\ \texttt{\{predicates\}}

\medskip

Now produce your refinement patch for the information below.

\medskip

\textbf{Task Information}

\begin{itemize}
    {\leftskip=1em
    \item Instruction: \\ \texttt{\{instruction\}} \medskip
    \item Objects in scene: \\ \texttt{\{objects\}} \medskip
    \par}
\end{itemize}
\medskip

\textbf{Action Information}

\begin{itemize}
    {\leftskip=1em
    \item Previously executed actions: \\ \texttt{\{executed\_actions\}} \medskip
    \item Erroneous action: \\ \texttt{\{erroneous\_action\}} \medskip
    \item Remaining actions (excluding erroneous action): \\ \texttt{\{remaining\_actions\}} \medskip
    \par}
\end{itemize}
\medskip

\textbf{Error Detail}

\begin{itemize}
    {\leftskip=1em
    \item State when error occurred: \\ \texttt{\{error\_state\}} \medskip
    \item Unfulfilled Preconditions: \\ \texttt{\{unfulfilled\_preconditions\}} \medskip
    \par}
\end{itemize}
\medskip

\textbf{Refinement rationale:}

\medskip

\textbf{Refinement patch:}

\end{tcolorbox}

\subsection{Common}
\begin{tcolorbox}[
    colback=lightgray,
    colframe=darkgray,
    coltitle=white,
    title=Common | State Prediction,
    fonttitle=\bfseries,
    arc=1mm,
    breakable
]

\textbf{\large \texttt{[System]}}

You are a state predictor for robotics.
The user will provide you with three synchronized images of the scene (top/front/back views) and partial ground truth object states.
Your task is to output state as list of grounded atoms which additionally holds true for the provided scene and information.

\medskip

\textbf{Output requirements:}

\begin{itemize}
    {\leftskip=1em
    \item Start the grounded atoms with unordered dashes (-).
    \item Only include atoms that are strongly supported by the images, output ``None" if no additional atoms can be inferred.
    \item Do not repeat information already provided in object state.
    \par}
\end{itemize}

\tcblower
\textbf{\large \texttt{[User]}}

Your task is to output state as list of grounded atoms which additionally holds true for the provided scene and information.

\medskip

Below are the definitions of the predicates. \\ \texttt{\{predicates\}}

\medskip

Below are examples. \\ \texttt{[Example]}

\medskip

Now provide the scene descriptions for the following sequences.

\medskip

\textbf{Information:}

\begin{itemize}
    {\leftskip=1em
    \item Instruction: \\ \texttt{\{instruction\}} \medskip
    \item Objects in scene: \\ \texttt{\{objects\}} \medskip
    \item Object state in scene: \\ \texttt{\{object\_state\}} \medskip
    \par}
\end{itemize}

\medskip

\textbf{State Predictions:}

\end{tcolorbox}

\begin{tcolorbox}[
    colback=lightgray,
    colframe=darkgray,
    coltitle=white,
    title=Common | Code Generation,
    fonttitle=\bfseries,
    arc=1mm,
    breakable
]

\textbf{\large \texttt{[System]}}

You are a python code generator for robotics. 
The user will provide two things:

\medskip

1) A set of imported Python modules and docstrings describing the available functions. \\
2) A specification containing: 
\begin{itemize}
    {\leftskip=1em
    \item Instruction: A natural language command for the robot
    \item Objects in scene: List of available objects
    \item Demonstration summary: A sequence of high-level action steps that serve as a high-level plan.
    \par}
\end{itemize}

\medskip

\textbf{Output requirements:}

\begin{itemize}
    {\leftskip=1em
    \item Use only the provided Python libraries and functions. Do not import new libraries, create new APIs, or change function signatures.
    \item Adhere strictly to the given docstrings. Call functions exactly as defined, with the allowed parameters.
    \item Return only the Python code, enclosed in triple backticks (```python ... ''').
    \item Treat the demonstration summary as a high-level plans; follow its sequence based on the Instruction and available objects.
    \item Do not add extra steps unless they are implied by the demonstration summary or the instruction.
    \par}
\end{itemize}

\tcblower
\textbf{\large \texttt{[User]}}

Your task is to write robot control scripts in Python code.
The Python code should be general and applicable to different robotics environments.

\medskip

Below are the imported Python libraries and functions that you can use, you can not import new libraries. \\ \texttt{[Libraries and Functions]}

\medskip

Below shows the docstrings for these imported library functions that you must follow. You can not add additional parameters to these functions. \\ \texttt{[API Docstring]}

\medskip

Below are examples. \\ \texttt{[Example]}

\medskip

Now generate Python code that follows the given specification.

\medskip

\textbf{Specification:}

\begin{itemize}
    {\leftskip=1em
    \item Instruction: \\ \texttt{\{instruction\}} \medskip
    \item Objects in scene: \\ \texttt{\{objects\}} \medskip
    \item Demonstration summary: \\ \texttt{\{demo\_summary\}} \medskip
    \par}
\end{itemize}

\medskip

\textbf{Generated Code:}

\end{tcolorbox}

\clearpage
\end{document}